\documentclass[10pt, letter, onecolumn]{arxiv}

\usepackage{kantlipsum, lipsum}
\usepackage{dm-colors}
\usepackage{amsmath}
\usepackage{pstricks, pst-node}
\usepackage{verbatim}
\usepackage{multirow}
\usepackage{longtable}
\usepackage{pdflscape}
\usepackage{scalerel}
\usepackage{booktabs}
\usepackage{enumitem}
\usepackage{xspace}
\usepackage{bm}
\usepackage{bbm}
\usepackage{mathtools}
\usepackage{soul}
\usepackage{epsfig}
\usepackage{graphicx}
\usepackage{amssymb}
\usepackage{colortbl}
\usepackage{csquotes}
\usepackage{setspace}
\usepackage{colortbl}
\usepackage{bbding}
\usepackage{threeparttable}
\usepackage{tabularx,ragged2e}
\usepackage{placeins}
\usepackage{bbding}
\definecolor{darkpastelgreen}{rgb}{0.13, 0.55, 0.13}
\definecolor{darkpastelred}{rgb}{0.55, 0.13, 0.13}
\definecolor{mygray}{rgb}{0.85, 0.85, 0.85}
\usepackage[hang,flushmargin]{footmisc}

\usepackage{nameref}
\usepackage{varioref}
\usepackage{amssymb}
\usepackage{pifont}
\usepackage{rotating}

\usepackage[pagebackref=false,breaklinks=false,%
            colorlinks=true,bookmarks=true,citecolor=ourdarkblue,%
            urlcolor=ourdarkblue,linkcolor=ourdarkblue]{hyperref}
\usepackage[noabbrev,capitalize]{cleveref}
\usepackage{etoc}
\usepackage{algorithm}
\usepackage{algpseudocode}

\graphicspath{{figures/}}

\title{\Large{Towards Evaluating and Building Versatile Large Language Models for Medicine}}

\author[1,2,$\ast$]{Chaoyi Wu} 
\author[1,2,$\ast$]{Pengcheng Qiu} 
\author[3]{Jinxin Liu}
\author[4]{Hongfei Gu}
\author[3]{Na Li}
\author[1,2]{\\ \vspace{0.1cm} Ya Zhang}
\author[1,2,$\dag$]{Yanfeng Wang} 
\author[1,2,$\dag$]{Weidi Xie}

\affil[1]{\normalsize Shanghai Jiao Tong University \hspace{1cm}}

\affil[2]{\normalsize Shanghai AI Laboratory  \authorcr \vspace{0.1cm}
}
\affil[3]{\normalsize China Mobile Communications Group Co., Ltd. \authorcr \vspace{0.1cm}
}
\affil[4]{\normalsize China Mobile Communications Group Shanghai Co., Ltd. \authorcr
}

\renewcommand{\correspondingauthor}[1]{$\ast$~Equal contributions. \\ $\dag$~Corresponding author. Email addresses:  \{wtzxxxwcy02, henrychur, weidi\}@sjtu.edu.cn}

\begin{document}

\begin{abstract}
In this study, we present \textbf{MedS-Bench}, a comprehensive benchmark designed to evaluate the performance of large language models (LLMs) in clinical contexts. Unlike existing benchmarks that focus on multiple-choice question answering, \textbf{MedS-Bench} spans 11 high-level clinical tasks, including clinical report summarization, treatment recommendations, diagnosis, named entity recognition, and medical concept explanation, among others. We evaluated six leading LLMs, {\em e.g.}, MEDITRON, Mistral, InternLM 2, Llama 3, GPT-4, and Claude-3.5 using few-shot prompting, and found that even the most sophisticated models struggle with these complex tasks.
To address these limitations, we developed \textbf{MedS-Ins}, a large-scale instruction tuning dataset for medicine. 
\textbf{MedS-Ins} comprises 58 medically oriented language corpora, totaling 5M instances with 19K instructions,
across 122 tasks. 
To demonstrate the dataset's utility, we conducted a proof-of-concept experiment by performing instruction tuning on a lightweight, open-source medical language model. The resulting model, \textbf{MMedIns-Llama 3}, significantly outperformed existing models across nearly all clinical tasks.
To promote further advancements in the application of LLMs to clinical challenges, 
we have made the \textbf{MedS-Ins} dataset fully accessible and invite the research community to contribute to its expansion.
Additionally, we have launched a dynamic leaderboard for \textbf{MedS-Bench}, which we plan to regularly update the test set to track progress and enhance the adaptation of general LLMs to the medical domain.
\textbf{Leaderboard}: \url{https://henrychur.github.io/MedS-Bench/}.
\textbf{Github}: \url{https://github.com/MAGIC-AI4Med/MedS-Ins}.

\end{abstract}

\maketitle


\section{Introduction}
Large Language Models (LLMs) have recently achieved significant advancements across various natural language processing tasks, demonstrating remarkable capabilities in language translation, text generation, dialogue, and beyond. These developments have also extended into the medical domain, 
where LLMs have achieved high scores on multiple-choice question-answering (MCQA) benchmarks in healthcare, and successfully passed the UMLS examination, as noted by Singhal et al.~\cite{singhal2023large, singhal2023towards}. Moreover, LLMs have shown expert-level performance in clinical text summarization when appropriate prompting strategies are employed~\cite{van2024adapted}.

Alongside these advancements, however, there has been growing criticisms and concerns regarding the application of LLMs in clinical settings, 
primarily due to their deficiencies in fundamental medical knowledge.
For instance, LLMs have demonstrated poor comprehension of ICD codes~\cite{soroush2024large}, produced inaccurate predictions related to clinical procedures~\cite{hager2024evaluation}, and misinterpreted Electronic Health Record (EHR) data~\cite{fleming2024medalign}. We posit that these polarized views on the efficacy of LLMs arise from the stringent standards required for AI deployment in clinical environments. 
Current benchmarks, which largely focus on multiple-choice problems~\cite{wu2024pmc,singhal2023towards,chen2023meditron}, fail to adequately reflect the practical utility of LLMs in real-world clinical scenarios.

To address this gap, we introduce \textbf{MedS-Bench} (S for Super), 
a comprehensive benchmark that extends beyond multiple-choice question answering~(MCQA), to include \textbf{11 advanced clinical tasks}, such as \textit{clinical report summarization}, \textit{treatment recommendations}, \textit{diagnosis}, and \textit{named entity recognition}, \textit{among others}. 
This benchmark provides clinicians and researchers with a detailed understanding of where LLMs excel and where they fall short in medical tasks. Specifically, we evaluate six mainstream models for medicine: 
MEDITRON~\cite{chen2023meditron}, Mistral~\cite{jiang2023mistral}, InternLM 2~\cite{cai2024internlm2}, Llama 3~\cite{touvron2023llama}, GPT-4~\cite{jpn-med-exam_gpt4} and Claude-3.5~\cite{claude}. Our findings indicate that even the most advanced LLMs struggle with complex clinical tasks, even when utilizing few-shot prompting, underscoring the gap between high performance on MCQA benchmarks and the actual demands of clinical practice.

To advance the development of open-source medical LLMs capable of tackling a broad spectrum of clinical tasks, we take inspiration from the idea of Super-NaturalInstructions~\cite{wang2022super}, 
and construct the first, comprehensive instruction tuning dataset for medicine, \textbf{MedS-Ins}.
It aggregates \textbf{58} open-source biomedical natural language processing datasets from five text sources, including exams, clinical texts, academic papers, medical knowledge bases, and daily conversations, 
resulting in \textbf{5M instances with 19K instructions} across \textbf{122 clinical tasks}, each accompanied with hand-written task definitions. We performed extensive instruction tuning on open-source medical language models, 
and explored both zero-shot and few-shot prompting strategies. The outcome is a new medical LLM -- \textbf{MMedIns-Llama 3}, for the first time, showing the effectiveness of training on diverse medical tasks through instruction tuning, enabling open-source medical LLMs to surpass leading closed-source models, including GPT-4 and Claude-3.5, across a wide range of clinical tasks.

While our final model serves primarily as an academic proof of concept, 
we believe that \textbf{MedS-Ins} represents an initial step toward advancing medical LLMs for real-world clinical applications, moving beyond the confines of online chat or multiple-choice question answering.

\section{Results}
In this section, we first introduce \textbf{MedS-Bench}, the benchmark employed in our study, designed to provide a comprehensive evaluation across a range of tasks critical to clinical applications. 
We then present detailed statistics on our instruction tuning dataset, \textbf{MedS-Ins}, which was carefully curated to cover a broad spectrum of medical language processing tasks. 
Finally, we provide an in-depth analysis of the evaluation results, 
comparing the performance of leading mainstream models with our own model, \textbf{MMedIns-Llama 3}, adapted from an open-source language model and fine-tuned on comprehensive medical instructions.

To ensure clarity in our subsequent discussion and analysis, we define key terminologies used throughout this study. 
For additional examples, please refer to Table~\ref{sec:detail MedSIns} in Supplementary.
\vspace{-0.2cm}
\begin{itemize}
    \setlength\itemsep{0.3em}
    \item \textbf{Text domains:} 
    Refers to the nature or type of the data, 
    such as clinical texts, examination materials, academic papers, and so forth.
    
    \item \textbf{Data sources:} 
    In contrast to the ``text domains'' which describe the attribute of the data, 
    ``data sources'' refer to the specific origins of the data, such as MIMIC-IV or PubMed papers. 
    Different data sources may belong to the same text domain.

    \item \textbf{Task categories:} 
    These denote the broad types of language processing tasks, such as multiple-choice question answering or named entity recognition, {\em etc}. Tasks within the same category share a common objective.
    
    \item \textbf{Tasks:} 
    These denote the fundamental units (leaf nodes) in our data collection pipeline, including specific tasks like outcome extraction, drug dose extraction, pathology summarization, {\em etc}. Each task may be defined by unique combinations of data sources, task categories, or text domains.
\end{itemize}

\subsection{The Description of MedS-Bench}
\begin{figure}[!t]
    \centering
    \includegraphics[width=\textwidth]{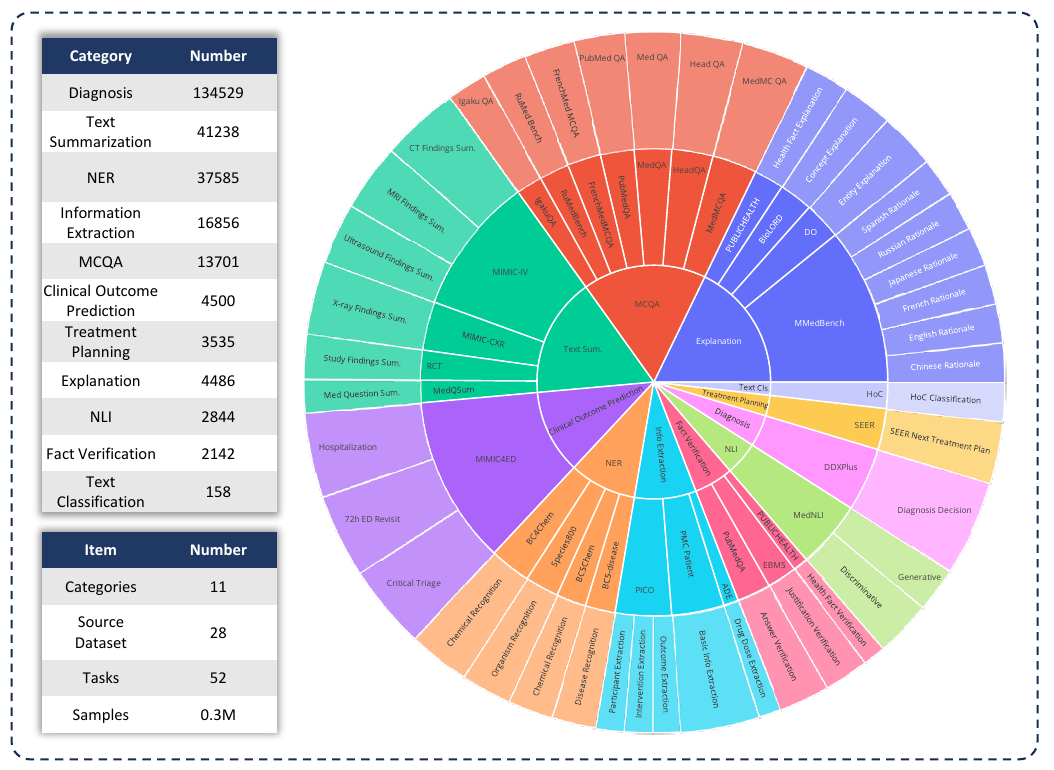}
    \caption{\textbf{Benchmark Statistics.} The hierarchical ring chart meticulously displays the data distribution within the evaluation benchmarks. The first tier categorizes the types of tasks, with the benchmarks encompassing 11 primary task categories. The second tier outlines the datasets involved, including 28
    datasets in total. The third tier details the specific tasks, with the benchmarks collectively addressing 52 distinct tasks. Overall, this benchmark allows for a thorough and comprehensive evaluation of model performance across multiple dimensions.}
    \label{fig:benchmark_statistics}
\end{figure}

\begin{figure}[!htbp]
    \centering
    \includegraphics[width=0.97\textwidth]{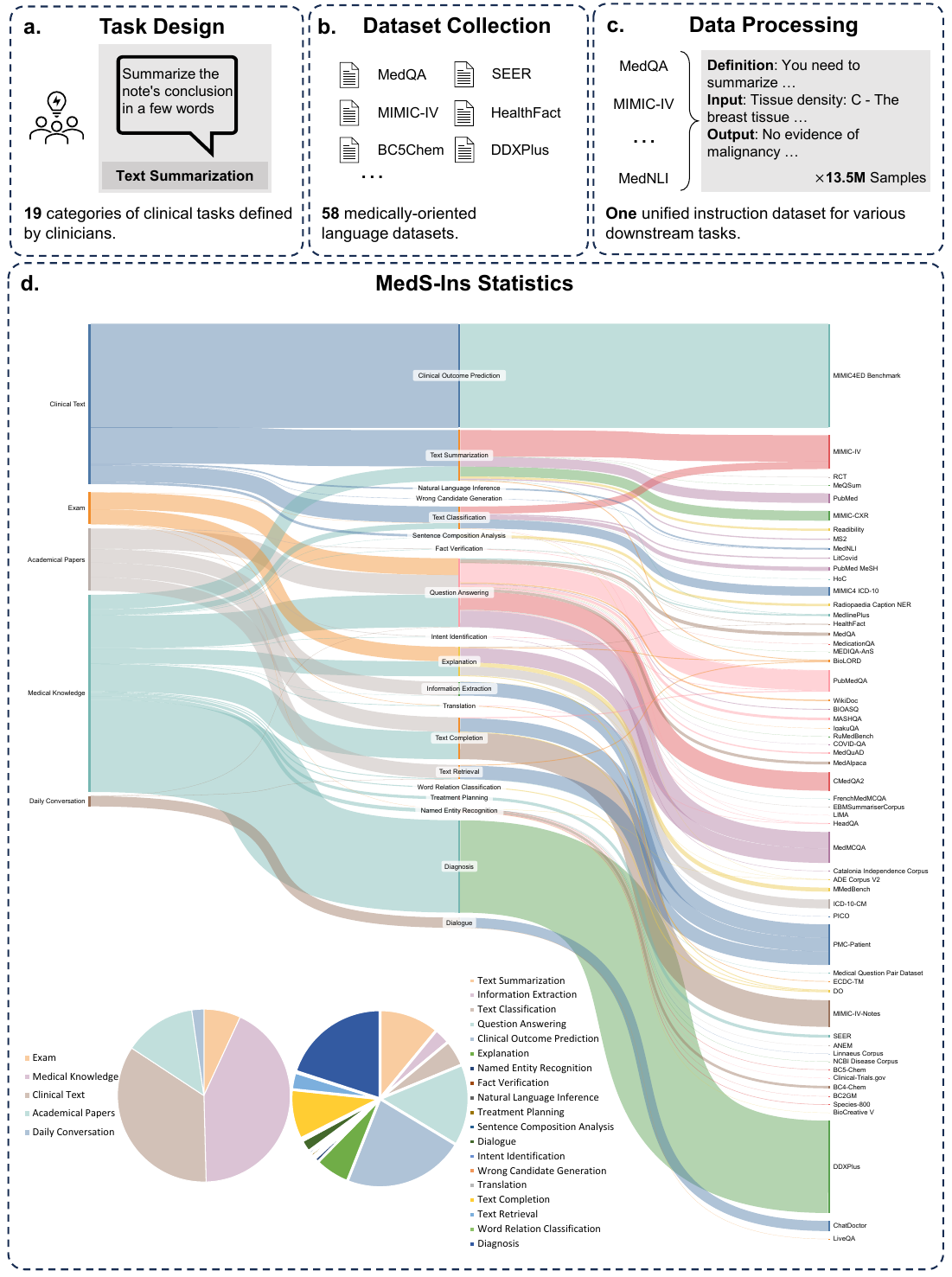}
    \caption{\textbf{Overview of MedS-Ins.} 
    \textbf{a} The task collection pipeline. For each task, we add a task category along with a hand-written definition to it, 
    resulting in a total of 19 task categories.  
    \textbf{b} We collect the existing 58 public datasets.
    \textbf{c} We convert the formats of different datasets into one unified medical instruction dataset, \textbf{MedS-Ins}. 
    \textbf{d} The final data distribution of our collected \textbf{MedS-Ins}. The Sankey diagram shows how the different text domains~(left), task categories~(middle), and data sources~(right) contribute to the final datasets. On the left of the bottom, two pie charts show the data distributions on text domains and task categories respectively.}
    \label{fig:topics_distribution}
\end{figure}

To evaluate the capabilities of various LLMs in clinical applications, 
we developed \textbf{MedS-Bench}, a comprehensive medical benchmark that extends beyond traditional multiple-choice questions. \textbf{MedS-Bench} encompasses \textbf{11 high-level clinical task categories}, derived from \textbf{28 existing datasets}, as illustrated in Figure~\ref{fig:benchmark_statistics}. Each dataset was reformatted into an instruction-prompted question-answering structure, complete with hand-crafted task definitions (instructions), as shown in Figure~\ref{fig:topics_distribution}a. The task categories we considered include: \textit{Multi-choice Question Answering, Text Summarization, Information Extraction, Explanation, Rationale, Named Entity Recognition, Diagnosis, Treatment Planning, Clinical Outcome Prediction, Text Classification, Fact Verification, and Natural Language Inference}. 
A more detailed description of each category is provided in Supplementary Sec.~\ref{Sec: Benchmark Details}.

In addition to defining these task categories, we also provide detailed statistics on the number of tokens and distinguish the required competencies for LLMs to address each task, as presented in Table~\ref{tab:benchmark_statistics} in Supplementary. Following previous work~\cite{jin2023cost}, we manually classified the tasks into two categories based on the skills required: (i) recalling facts from the model, and (ii) retrieving facts from the provided context. 
Broadly speaking, the former involves tasks that require to access knowledge encoded in the model's weights from large-scale pre-training, while the latter involves tasks that necessitate extracting information from the provided context, such as in summarization or information extraction. As shown in Table~\ref{tab:benchmark_statistics} in Supplementary, eight of the task categories require the model to recall knowledge from the model, while the remaining three require fact retrieval from the given context.

\subsection{The Description of MedS-Ins}
In this section, we introduce our proposed instruction dataset, \textbf{MedS-Ins}, 
with the data collected from 5 distinct text sources and 19 task categories, 122 distinct clinical tasks. 
The statistics of \textbf{MedS-Ins} are summarized in Figure~\ref{fig:topics_distribution}.

\subsubsection*{Text Domains}
Our proposed instruction tuning dataset is composed of samples drawn from five distinct sources: exams, clinical texts, academic papers, medical knowledge bases, and daily conversations.

    \noindent \textbf{Exams:} 
    This category consists of data from medical examination questions across various countries. 
    It encompasses a broad spectrum of medical knowledge, ranging from fundamental medical facts to complex clinical procedures. While the exam domain is a vital resource for understanding and assessing medical education, 
    it is important to note that the highly standardized nature of exams often results in over-simplified cases compared to real-world clinical tasks. 7\% of the tokens in our dataset are from the exams. 

    \noindent \textbf{Clinical texts:} 
    Generated during routine clinical practice, these texts support diagnostic, treatment, and preventive processes within hospitals and clinical centers. This category includes Electronic Health Records (EHRs), radiology reports, lab results, follow-up instructions, and medication recommendations, among others. These texts are indispensable for disease diagnosis and patient management, making accurate analysis and understanding crucial for the effective clinical application of LLMs. 35\% of the tokens in our dataset are from clinical texts. Notably, the significant proportion of clinical texts ensures that the instruction tuning data aligns closely with clinical demands. 

    \noindent \textbf{Academic papers:}
    This data is sourced from medical research papers, covering the latest findings and advancements in the medical research field. Given their accessibility and structured organization, extracting data from academic papers is relatively straightforward. These cases help models grasp cutting-edge medical research information, guiding them to better understand contemporary developments in medicine. 13\% of the tokens in our dataset are from academic papers.
     
    \noindent \textbf{Medical knowledge bases:} 
    This domain comprises well-organized and comprehensive medical knowledge, including medical encyclopedias, knowledge graphs, and glossaries of medical terms. Such data forms the backbone of medical knowledge bases, supporting both medical education and the application of LLMs in clinical practice. 43\% of the tokens in our dataset are from medical knowledge.
    
    \noindent \textbf{Daily conversations:} 
    This source refers to the daily consultation generated between doctors and patients, 
    primarily sourced from online platforms and other interactive scenarios. 
    These interactions reflect the real-life interactions between medical professionals and patients, 
    playing a critical role in understanding patients' needs and enhancing the overall experience of medical service.
    2\% of the tokens in our dataset are from daily conversations.

\subsubsection*{Task Categories}
\label{categories}

Beyond categorizing the text domains from which the original data is sourced, 
the samples in \textbf{MedS-Ins} are further organized into distinct task categories. 
We have identified 19 task categories, each representing a critical capability that we believe a medical LLM should possess. By constructing this instruction-tuning dataset and fine-tuning models accordingly, we aim to equip the LLMs with the versatility needed to address a broad spectrum of medical applications.

These 19 task categories include but are not limited to, the 11 categories in the MedS-Bench benchmark. 
The additional categories encompass a range of linguistic and analytical tasks essential for comprehensive medical language processing, including \textit{Intent Identification, Translation, Word Relation Classification, Text Retrieval, Sentence Composition Analysis, Wrong Candidate Generation, Dialogue}, and \textit{Text Completion} and the MCQA category is extended to general \textit{Question Answering}, which also includes free-text answering cases.
The diversity of task categories—ranging from common question answering and dialogue to various downstream clinical tasks—guarantees a comprehensive understanding of the potential medical applications.
A detailed description of each category is provided in Supplementary Sec.~\ref{Sec: Benchmark Details}. 

\subsection{Quantitative Results on Various Tasks}
For each task type, we start by discussing the performance of various existing LLMs, followed by a comparison with our final model, \textbf{MMedIns-Llama 3}. 
All results presented here were obtained using a 3-shot prompting strategy~(more details in Supplementary Sec.~\ref{Sec:Evaluation Settings}), except for MCQA tasks, where we used a zero-shot prompting setting to align with previous studies~\cite{chen2023meditron,wu2024pmc,qiu2024towards}. 
As our comparisons include proprietary models like GPT-4 and Claude 3.5, 
which incur usage costs, we randomly sampled around 1500 test cases per benchmark to manage the cost constraints. The task description and specific sampling numbers are detailed in Supplementary Sec.~\ref{Sec: Benchmark Details}. For simplicity, the percentage mark (\%) is omitted in all the following tables and results analysis.


\subsubsection*{Multilingual Multiple-choice Question-answering}
\label{sec:Results}
Here, we present evaluation results on the widely used multiple-choice Question-answering~(MCQA) benchmarks, as shown in Table~\ref{tab:acc_benchmark_comparison}. 
Some numbers are directly incorporated from our previous studies~\cite{qiu2024towards,wu2024pmc}. 

On these multi-choice question answering datasets, existing proprietary LLMs have demonstrated very high accuracies, for example, on MedQA, GPT-4 can achieve 85.8, which is almost comparable to human experts, and Llama 3 can also pass the exam with 60.9 scores. Similarly, in languages other than English, 
LLMs also demonstrate superior results in multiple-choice accuracy on MMedBench~\cite{qiu2024towards}. The results indicate that as multi-choice questions have been extensively considered in existing research, different LLMs may have been specifically optimized for such tasks, resulting in high performance. 
It is therefore essential to build up a more comprehensive benchmark, 
to further push the development of LLMs towards clinical applications.

Our proposed model, {MMedIns-Llama 3}, although not primarily trained on multi-choice questions, still shows notable improvement, achieving an average accuracy of 63.9 across different benchmarks, significantly surpassing GPT-3.5.

\begin{table}[!htb]
\centering
\begin{threeparttable}[]
\footnotesize
\tabcolsep=0.32cm
    \caption{Results on medical multiple-choice question answering, as reported with Accuracy score. Bolding represents the best results.}
    \label{tab:acc_benchmark_comparison}
   \begin{tabular}{l|cccccccc|c}
    \toprule
    \multirow{2}{*}{Method} & \multirow{2}{*}{MedQA} & \multirow{2}{*}{MedMCQA} & \multirow{2}{*}{PubMedQA} & \multicolumn{5}{c|}{MMedBench}    & \multirow{2}{*}{Avg.} \\ \cline{5-9}
                        &  &                          &                           & ZH & JA & FR & RU & ES &                        \\
    \midrule
    \rowcolor{mygray} \multicolumn{10}{c}{Close-source Models} \\ 
    \midrule
    GPT-3.5 & 57.7 & \textbf{72.7} & 53.8 & 52.3 & 34.6 & 32.5 & 66.4 & 66.1 & 54.5\\ 
    GPT-4   & \textbf{85.8} & 72.3 & 75.2 & 75.1 & \textbf{72.9} & \textbf{56.6} & \textbf{83.6} & \textbf{85.7} & \textbf{75.3} \\ 
    \rowcolor{mygray} \multicolumn{10}{c}{Open-source Models} \\ 
    \midrule
    MEDITRON & 47.9 & 59.2 & 74.4 & 61.9 & 40.2 & 35.1 & 67.6 & 53.3 & 54.9 \\ 
    InternLM 2 & - & - & - & 77.6 & 47.7 & 41.0 & 68.4 & 59.6 & - \\ 
    Mistral & 50.8 & 48.2 & 75.4 & 71.1 & 44.7 & 48.7 & 74.2 & 63.9 & 49.1 \\ 
    Llama 3 & 60.9 & 50.7 & 73.0 & 78.2 & 48.2 & 50.8 & 71.5 & 64.2 & 62.2 \\ 
    \midrule
    MMedIns-Llama 3 & 63.6 & 57.1 & \textbf{78.2} & \textbf{78.6} & 54.3 & 46.0 & 72.3 & 61.2 & 63.9 \\ 
    \bottomrule
    \end{tabular}
  \end{threeparttable}
  \vspace{4pt}
\end{table}

\subsubsection*{Text Summarization}

As shown by Table~\ref{TextSummarizationResults}, the performance of text summarization is reported as `BLEU/ROUGE' scores, on multiple report types across various modalities, including X-ray, CT, MRI, ultrasound, and other medical questions. 
Among the models, closed-source LLMs, such as GPT-4 and Claude-3.5, 
perform better than all the open-source ones, achieving an average of 24.46/25.66 and 26.29/27.36, respectively. In open-source models, Mistral\ 3 demonstrates the best results, with 24.48/24.90 in BLEU/ROUGE respectively and Llama 3 follows, with 22.20/23.08.

Our model ({MMedIns-Llama 3}), trained on the medical-specific instruction dataset (\textbf{MedS-Ins}), significantly outperforms the others, including the top closed-source model, GPT-4 or Claude-3.5, achieving average scores of 46.82/48.38.

\begin{table}[!htb]
\centering
\begin{threeparttable}[]
\footnotesize
\caption{Results on text summarization, as reported with BLEU and ROUGE scores, formatted as `BLEU/ROUGE'. ``Con.'' denotes conclusions. Bolding represents the best results.}
\label{TextSummarizationResults}
\tabcolsep=0.12cm
   \begin{tabular}{l|cccccc|c}
    \toprule
        \multirow{2}{*}{Method} & MedQSum & RCT-Text & MIMIC-CXR & \multicolumn{3}{c|}{MIMIC-IV} & \multirow{2}{*}{Avg.} \\
        \cline{5-7}
         & Med Question & Study Con. & X-ray & Ultrasound & CT & MRI &   \\ \hline
        \rowcolor{mygray} \multicolumn{8}{c}{Close-source Models} \\
        GPT-4 & 25.06/27.30 & 34.32/31.09 & 27.26/29.71 & 11.17/14.53 & 23.97/29.52 & 25.76/32.06 & 24.46/25.66 \\ 
        Claude-3.5 & 21.14/25.06 & 41.02/36.16 & 27.76/29.93 & 15.24/18.28 & 21.98/26.38 & 26.43/31.05 & 26.29/27.36 \\ \midrule
        \rowcolor{mygray} \multicolumn{8}{c}{Open-source Models} \\
        MEDITRON & 15.64/23.14 & 4.00/16.44 & 5.21/16.50 & 3.75/6.07 & 16.30/23.93 & 20.11/27.98 & 7.15/15.54 \\ 
        InternLM 2 & 15.69/21.63 & 14.48/15.16 & 11.83/13.41 & 13.48/20.96 & 20.88/27.82 & 23.43/31.40 & 13.87/17.79 \\ 
        Mistral & 23.49/26.03 & 27.24/26.13 & 22.09/24.71 & 25.09/22.72 & 27.60/30.77 & 29.87/31.81 & 24.48/24.90 \\ 
        Llama 3 & 22.45/25.08 & 15.38/14.60 & 32.92/32.64 & 18.06/20.00 & 24.47/29.35 & 24.82/30.50 & 22.20/23.08 \\  \midrule
        MMedIns-Llama 3 & \textbf{54.16/56.95} & \textbf{57.82/55.60} & \textbf{54.91/57.64} & \textbf{20.40/23.32} & \textbf{42.18/46.46} & \textbf{40.53/43.38} & \textbf{46.82/48.38} \\ 
    \bottomrule
    \end{tabular}
  \end{threeparttable}
  \vspace{4pt}
\end{table}

\subsubsection*{Information Extraction}
The performance of information extraction is summarized in Table~\ref{InformationExtractionResults}. 
InternLM 2 shows exceptionally good performance in this task with an average score of 79.11. For example, In the PICO tasks, InternLM 2 leads in both Intervention and Outcome Extraction, with scores of 74.42 and 69.77, respectively. Closed-source models such as GPT-4 and Claude-3.5 outperform all other open-source counterparts, with average scores of 76.92 and 79.41, respectively. Analysis of individual benchmark components reveals that most LLMs perform better at extracting less medically complex information, such as basic patient details, compared to more specialized medical data like outcomes and interventions. For instance, in extracting basic information from PMC Patients, most LLMs score above 90, with Claude-3.5 achieving the highest score of 99.07. In contrast, performance on Clinical Outcome Extraction tasks within PICO is relatively poor.

Our proposed model, {MMedIns-Llama 3}, demonstrates the best overall performance, achieving an average score of 83.77, surpassing InternLM 2 by 4.66 points. Notably, in the PICO tasks, MMedIns-Llama 3 excels in Participant Extraction, scoring 83.72, which exceeds the second-best model by 11.63 points.

\begin{table}[!htb]
\centering
\begin{threeparttable}[]
\footnotesize
\caption{Results on information extraction, as reported with Accuracy score.  
``Ext.'' denotes extraction and ``Info.'' denotes information. Bolding represents the best results. }
\label{InformationExtractionResults}
\tabcolsep=0.22cm
   \begin{tabular}{l|ccccc|c}
    \toprule
        \multirow{2}{*}{Method} & \multicolumn{3}{c}{PICO} & \multicolumn{1}{|c|}{ADE} & \multicolumn{1}{c|}{PMC patient}  & \multirow{2}{*}{Avg.}\\
        \cline{2-4}
         & Participant Ext. & Intervention Ext. & Outcome Ext. & \multicolumn{1}{|c|}{Drug Dose Ext.} & \multicolumn{1}{c|}{Basic Info. Ext.}   &   \\
    \hline
    \rowcolor{mygray} \multicolumn{7}{c}{Close-source Models} \\
        GPT-4 & 67.44 & 62.79 & 65.12 & 91.30 & 97.93 & 76.92 \\ 
        Claude-3.5 & 65.12 & 76.74 & 60.47 & 95.65 & \textbf{99.07} & 79.41 \\ 
        \midrule
     \rowcolor{mygray} \multicolumn{7}{c}{Open-source Models} \\
        MEDITRON & 72.09 & 46.51 & 51.16 & \textbf{95.65} & 72.20 & 67.52 \\ 
        InternLM 2 & 72.09 & 74.42 & \textbf{69.77} & \textbf{95.65} & 83.60 & 79.11 \\ 
        Llama 3 & 58.14 & 79.07 & 58.14 & 69.57 & 95.93 & 72.17 \\ 
        Mistral & 60.47 & 65.12 & 48.84 & 91.30 & 85.20 & 70.18 \\ 
        \midrule
        MMedIns-Llama 3 & \textbf{83.72} & \textbf{79.07} & 62.79 & \textbf{95.65} & 97.60 & \textbf{83.77} \\ 
    \bottomrule
    \end{tabular}
  \end{threeparttable}
  \vspace{4pt}
\end{table}

\subsubsection*{Concept Explanation}

We conduct evaluations on medical concept explanation and reported the BLEU-1 and ROUGE-1 scores across all relevant datasets and models.

In Table~\ref{ExplanationTaskResults}, we evaluate the model on medical concept explanation, GPT-4, Llama 3, and Mistral perform well on this task, 
achieving an average of 19.37/21.58, 13.51/17.92, and 13.53/17.37 respectively. On the contrary, the scores of Claude-3.5, InternLM 2, and MEDITRON are relatively lower, with averages of 12.56/16.75, 11.53/17.01, and 8.51/18.90, respectively. We conjecture that the relatively poor performance of MEDITRON can be attributed to its training corpus, which is more focused on academic papers and guidelines, and thus falls short in basic medical concept explanation.

Our final model, {MMedIns-Llama 3}, significantly outperforms the other ones across all concept explanation tasks, particularly in Health Fact Explanation (30.50/28.53) and BioLORD Concept Explanation (38.12/43.90),
and achieving the highest average scores of 34.43/37.47. Following MMedIns-Llama 3, GPT-4 also showed strong performance, with GPT-4 scoring 19.37/21.58.

\begin{table}[!htb]
\centering
\begin{threeparttable}[]
\footnotesize
\tabcolsep=0.5cm
    \caption{Results on medical concept explanation, as reported with `BLEU/ROUGE' scores. ``Exp.'' denotes Explanation. Bolding represents the best results.}
    \label{ExplanationTaskResults}
   \begin{tabular}{l|ccc|c}
    \toprule
        Method & Health Fact Exp. & Do Entity Exp. & BioLORD Concept Exp. &  Avg. \\ 
    \hline
    \rowcolor{mygray} \multicolumn{5}{c}{Close-source Models} \\
        GPT-4 & 18.63/20.80 & 19.14/21.14 & 20.33/22.80 & 19.37/21.58 \\ 
        Claude-3.5 & 14.96/18.48 & 8.75/13.28 & 13.95/18.49 & 12.56/16.75 \\ 
        \midrule
        \rowcolor{mygray} \multicolumn{5}{c}{Open-source Models} \\
        MEDITRON & 6.09/8.65 & 7.68/25.39 & 11.76/22.66 & 8.51/18.90 \\ 
        InternLM 2 & 22.36/27.01 & 5.28/10.39 & 6.95/13.62 & 11.53/17.01 \\ 
        Llama 3 & 16.79/20.32 & 14.88/18.84 & 8.87/14.61 & 13.51/17.92 \\ 
        Mistral & 18.11/21.31 & 9.21/14.11 & 13.27/16.68 & 13.53/17.37 \\ 
        \midrule
        MMedIns-Llama 3 & \textbf{30.50/28.53} & \textbf{34.66/39.99} & \textbf{38.12/43.90} & \textbf{34.43/37.47} \\ 
    \bottomrule
    \end{tabular}
  \end{threeparttable}
  \vspace{4pt}
\end{table}

\subsubsection*{Answer Explanations~(Rationale)}
In Table~\ref{ReasoningTaskResults}, we evaluate the complex rationale,
{\em i.e.}, explaining the answer and comparing the reasoning abilities of various models using the \textbf{MMedBench}~\cite{qiu2024towards} dataset across six languages. Among the models tested, the closed-source model Claude-3.5 exhibited the strongest performance, with average scores of 46.26/36.97, demonstrating consistently high scores across all languages, particularly in French and Spanish. This superior performance may be attributed to the similarity of this task to chain-of-thought reasoning, a capability that has been specifically enhanced in many general-purpose LLMs. 
Among open-source models, Mistral and InternLM 2 showed comparable performance, with average scores of 38.14/32.28 and 35.65/32.04, respectively.
It is important to note that GPT-4 was excluded from this evaluation, 
because the rationale component of the MMedBench dataset was primarily constructed using GPT-4 outputs, which could introduce bias and bring unfair comparisons.

Consistent with our observations in concept explanation, our final model, {MMedIns-Llama 3}, demonstrated the best overall performance 
with average BLEU-1/ROUGE-1 scores of 46.90/34.54 across all languages,
notably, achieving 51.74/35.19 in Japanese reasoning tasks, 
49.08/38.19 in English, and 46.93/38.73 in French, respectively.
This superior performance is likely due to the fact that our base language model~(MMed-Llama 3) was initially developed to be multilingual~\cite{qiu2024towards}. Consequently, even though our instruction tuning did not explicitly target multilingual data, the final model outperforms others across multiple languages. 

\begin{table}[!htb]
\centering
\begin{threeparttable}[]
\footnotesize
\tabcolsep=0.14cm
    \caption{Results on rationale, as reported with `BLEU/ROUGE' scores.
    Note here, we do not include the results for GPT-4 since the original evaluation sets were generated with it, that may bring unfair comparison bias. Bolding represents the best results.}
    \label{ReasoningTaskResults}
   \begin{tabular}{l|cccccc|c}
    \toprule
        \multirow{2}{*}{Method}  & \multicolumn{6}{c|}{MMedBench} &  \multirow{2}{*}{Avg.}\\ 
        \cline{2-7}
        &Chinese&English&French&Japanese&Russian&Spanish& \\
    \hline
        \rowcolor{mygray} \multicolumn{8}{c}{Close-source Models} \\ 
        Claude-3.5 & 44.64/34.63 & 47.07/38.67 & \textbf{48.93/41.23} & 49.22/\textbf{39.15} & \textbf{38.90/28.17} & \textbf{48.80/39.99} & 46.26/\textbf{36.97} \\ 
        \midrule
        \rowcolor{mygray} \multicolumn{8}{c}{Open-source Models} \\ 
        MEDITRON & 20.39/21.79 & 38.42/31.24 & 34.43/29.33 & 18.89/24.98 & 24.32/16.77 & 37.64/31.01 & 29.01/25.86 \\ 
        InternLM 2 & 35.23/30.77 & 44.12/37.39 & 36.10/33.65 & 29.13/33.15 & 27.43/20.99 & 41.87/36.30 & 35.65/32.04 \\ 
        Llama 3 & 28.51/23.30 & 44.10/39.26 & 24.92/22.24 & 13.46/15.04 & 31.16/22.85 & 32.37/27.70 & 29.09/25.06 \\ 
        Mistral & 35.53/28.91 & 47.20/37.88 & 39.53/35.64 & 29.16/28.96 & 32.15/23.99 & 45.27/38.33 & 38.14/32.28 \\ 
        \midrule
        MMedIns-Llama 3 & \textbf{50.27/34.01} & \textbf{49.08/38.19} & 46.93/38.73 & \textbf{51.74}/35.19 & 35.27/23.81 & 48.15/37.35 & \textbf{46.90}/34.54 \\ 
    \bottomrule
    \end{tabular}
  \end{threeparttable}
  \vspace{4pt}
\end{table}

\subsubsection*{Named Entity Recognition~(NER)}
As shown in Table \ref{HardNERTaskResults}, among the six existing models, 
GPT-4 is the only one that consistently demonstrates robust performance across Named Entity Recognition (NER) tasks, achieving an average F1-Score of 59.52. 
It excels particularly in the BC5Chem Chemical Recognition task with a score of 67.62. InternLM 2 also delivers competitive results, with an average F1-Score of 45.69, showing strong performance in both the BC5Chem and BC5Disease tasks. Llama 3 and Mistral, with average F1-Scores of 23.62 and 16.53, respectively, exhibit moderate performance. MEDITRON, not optimized for NER tasks, shows limited effectiveness in this area.

Our model, \textbf{MMedIns-Llama 3}, significantly outperforms all other models, achieving an an average F1-Score of 79.29. It excels in the BC4Chem and BC5Chem Chemical Recognition tasks, with F1 scores of 90.78 and 91.25, respectively. Furthermore, MMedIns-Llama 3 leads in the BC5Disease Disease Recognition task with an F1-Score of 54.26 and in the Species800 Organism Recognition task with 80.87, demonstrating superior capability in handling complex NER tasks across various biomedical domains.

\begin{table}[!htb]
\centering
\begin{threeparttable}[]
\footnotesize
\tabcolsep=0.55cm
    \caption{Results on NER tasks, as reported with F1-Score scores. 
    `Rec.' is short for `recognition'. Bolding represents the best results.}
    \label{HardNERTaskResults}
   \begin{tabular}{l|cccc|c}
    \toprule
        \multirow{2}{*}{Method} & \multicolumn{1}{c}{BC4Chem}& \multicolumn{1}{c|}{BC5Chem}  & \multicolumn{1}{c|}{BC5Disease} & \multicolumn{1}{c|}{Species800} &  \multirow{2}{*}{Avg.} \\ 
        \cline{2-3}
         & \multicolumn{2}{c|}{Chemical Rec.} & \multicolumn{1}{c|}{ Disease Rec.} & \multicolumn{1}{c|}{Organism Rec.} &   \\ 
    \hline
        \rowcolor{mygray} \multicolumn{6}{c}{Close-source Models} \\ 
        GPT-4 & 54.84 & 67.62 & 53.20 & 62.43 & 59.52 \\ 
        Claude-3.5 & 22.98 & 40.77 & 24.05 & 14.45 & 25.56 \\ 
        \midrule
        \rowcolor{mygray} \multicolumn{6}{c}{Open-source Models} \\ 
        MEDITRON & 1.98 & 4.11 & 1.33 & 0.40 & 1.96 \\ 
        InternLM 2 & 41.21 & 41.51 & 37.11 & 62.93 & 45.69 \\ 
        Llama 3 & 19.45 & 37.83 & 25.30 & 11.90 & 23.62 \\ 
        Mistral & 15.56 & 32.09 & 12.17 & 6.31 & 16.53 \\ 
        \midrule 
        MMedIns-Llama 3 & \textbf{90.78} & \textbf{91.25} & \textbf{54.26} & \textbf{80.87} & \textbf{79.29} \\ 
    \bottomrule
    \end{tabular}
  \end{threeparttable}
  \vspace{4pt}
\end{table}


\subsubsection*{Diagnosis, Treatment Planning, and Clinical Outcome Prediction}

We evaluate the performance on tasks involving diagnosis, 
treatment planning, and clinical outcome prediction, 
using the DDXPlus benchmark for Diagnosis, the SEER benchmark for Treatment Planning, and the MIMIC4ED benchmark for Clinical Outcome Prediction. 
The results, presented in Table~\ref{TreatmentPlanningResults}, are measured in terms of accuracy. Here, the use of accuracy as a metric is appropriate in this generative prediction problem, as each of these datasets simplifies the original problem into a closed-set choice. Specifically, DDXPlus utilizes a predefined list of diseases, from which models must select one based on the provided patient context. In SEER, treatment recommendations are categorized into eight high-level categories, while in MIMIC4ED, the final clinical outcome decisions are binary, with options of either True or False.

Overall, the open-source LLMs underperform the closed-source counterparts in these tasks, and in some instances, they fail to provide meaningful predictions. For example, Llama 3 struggles with predicting critical triage and the 72-hour ED revisit binary indicator. In the SEER treatment planning task, InternLM 2 and MEDITRON achieved moderate accuracy scores of 62.33 and 68.27, respectively. 
For the DDXPlus diagnosis task, InternLM 2 and Mistral perform slightly better, with an accuracy of 35.20 and 34.80 respectively. Notably, for clinical outcome prediction, we find Llama 3 fails in this task. It appears that because this task is outside its training distribution, it often fails to follow the three-shot provided format, thus resulting in extremely low scores, 9.27 and 8.80 in predicting 72h ED Revisit and Critical Triage respectively.
However, closed-source models such as GPT-4 and Claude-3.5 demonstrate significantly better performance. Claude-3.5, for instance, achieves a 92.93 accuracy score in treatment planning and GPT-4 attains 84.73. They also demonstrate better performance in diagnosis, highlighting the considerable gap between open-source and closed-source LLMs. Despite these results, the scores remain insufficient for reliable clinical use. 

In contrast, \textbf{MMedIns-Llama 3} demonstrate superior accuracy in clinical decision support tasks, with a 98.47 accuracy score in treatment planning, 97.53 in diagnosis, and an average accuracy of 63.35~(mean on the scores of Hospitalization, 72h ED Revisit, and Critical Triage) in clinical outcome prediction. 

\begin{table}[!htb]
\centering
\begin{threeparttable}[]
\footnotesize
\tabcolsep=0.17cm
\caption{Results on treatment planning, diagnosis clinical outcome prediction, and text classification results. The first 3 tasks are reported with Accuracy score, and text classification is reported with Precision, Recall, and F1 scores. Bolding represents the best results.}
   \label{TreatmentPlanningResults}
   \begin{tabular}{l|c|c|ccc|ccc}
    \toprule
        \multirow{2}{*}{Method} & \multirow{2}{*}{SEER} & \multirow{2}{*}{DDXPlus} & \multicolumn{3}{c|}{MIMIC4ED} & \multicolumn{3}{c}{HoC Classification} \\ 
        \cline{4-9}
        & & &Hospitalization &72h ED Revisit &Critical Triage  & Precision & Recall & F1 \\ \midrule
        \rowcolor{mygray} \multicolumn{9}{c}{Close-source Models} \\ 
        GPT-4 & 84.73       & 58.13  & 61.20 & \textbf{58.07} & 60.13  & 61.07 & 80.23 & 68.06\\ 
        Claude-3.5 & 92.93  & 60.24  & 65.80 & 57.91 & \textbf{68.53}   & 58.43 & 79.84 & 66.74\\ 
        \midrule
        \rowcolor{mygray} \multicolumn{9}{c}{Open-source Models} \\ 
        MEDITRON & 68.27    & 29.53  & 56.27 & 48.47 & 45.67   & 19.61 & 34.61 & 23.70\\ 
        InternLM 2 & 62.33  & 35.20  & 58.80 & 55.13 & 52.80  & 20.65 & 82.24 & 31.09 \\ 
        Llama 3 & 56.07     & 33.73  & 39.07 & 9.27 & 8.80    & 32.40 & 52.03 & 38.37 \\ 
        Mistral & 38.93     & 34.80  & 56.27 & 48.47 & 45.67   & 40.39 & 64.11 & 48.73\\ 
        \midrule
        MMedIns-Llama 3&\textbf{98.47}& \textbf{97.53} & \textbf{74.20} & 52.73 & 63.13  & \textbf{89.59} & \textbf{85.58} & \textbf{86.66}\\
    \bottomrule
    \end{tabular}
  \end{threeparttable}
  \vspace{4pt}
\end{table}

\subsubsection*{Text Classification}

In Table~\ref{TreatmentPlanningResults}, we present the evaluation on the Hallmarks of Cancer (HoC) multi-label classification task, and report macro-Precision, macro-Recall, and macro-F1 scores. For this task, all candidate labels are input into the language model as a list, and the model is asked to select its preferred answers, allowing for multiple selections. 
The metrics are then calculated based on these model selections.

GPT-4 and Claude-3.5 perform well on this task, with GPT-4 achieving a macro-F1 score of 68.06 and Claude-3.5 slightly worse at 66.74. Both models show strong recall capabilities, particularly GPT-4, which achieves a macro-Recall of 80.23, underscoring its proficiency in identifying relevant labels. 
Mistral shows moderate performance with a macro-F1 score of 48.73, 
maintaining a balance between precision and recall. 
In contrast, Llama 3 and InternLM 2 display lower overall performance, with macro-F1 scores of 38.37 and 31.09, respectively. These models, particularly InternLM 2, demonstrate high recall but struggle with precision, resulting in lower F1 scores. MEDITRON ranks lowest in this task, with a macro-F1 score of 23.70.

\textbf{MMedIns-Llama 3} clearly outperforms all other models, achieving the highest scores across all metrics, with a macro-Precision of 89.59, a macro-Recall of 85.58, and a macro-F1 score of 86.66. 
These results highlight MMedIns-Llama 3's superior ability to accurately classify and recall multiple labels, making it the most effective model for this complex task.

\subsubsection*{Fact Verification}
In Table~\ref{FactVerificationResults}, we evaluate the models on fact verification tasks. For PubMedQA Answer Verification and HealthFact Verification, the LLMs are required to select a single answer from a list of provided candidates, with accuracy serving as the evaluation metric. 
In contrast, for EBMS Justification Verification, where the task involves generating free-form text, performance is assessed using BLEU and ROUGE scores.

InternLM 2 achieves the highest accuracy on PubMedQA Answer Verification with scores of 99.23. On PUBLICHEALTH, it also achieves 76.94 accuracy scores, which is the best among all the open-sources models, just behind GPT-4 with 78.60. 
In the EBMS benchmark, GPT-4 demonstrates the strongest performance, with BLEU/ROUGE scores of 16.28/16.27.  
Llama 3 achieves an accuracy of 94.77 and 63.89 on PubMedQA and HealthFact Verification, respectively, and a BLEU/ROUGE score of 16.52/16.49 in the EBMS benchmark. Mistral, while achieving a high BLEU/ROUGE score of 15.98/16.43 in EMBS, has a relatively lower accuracy of 57.38 and 69.78. MEDITRON ranks lowest across the accuracy metrics, with accuracy of 25.23 and 32.66 on the PubMedQA and PUBLICHEALTH datasets. In EMBS, MEDITRON achieves a BLEU/ROUGE score of 11.58/15.78.

\textbf{MMedIns-Llama 3} continues surpassing existing models, achieving the highest accuracy score as InternLM 2, excelling in PubMedQA Answer Verification and HealthFact Verification while in EMBS, MMedIns-Llama 3 slightly falls behind the GPT-4 and Llama\ 3 with 12.71/14.65 in BLEU and ROUGE, which we treat as future work for further improvement.

\begin{table}[!htb]
\centering
\begin{threeparttable}[]
\footnotesize
\tabcolsep=0.26cm
    \caption{Results on fact verification and NLI results, 
    as reported with both accuracy and BLUE/ROUGE scores. 
    `Ver.' denotes `verification'. Bolding represents the best results.}
    \label{FactVerificationResults}
   \begin{tabular}{l|cc|c|cc}
    \toprule
        \multirow{2}{*}{Method}      & PubMedQA & PUBLICHEALTH  & EMBS & \multicolumn{2}{c}{MedNLI textual entailment} \\ 
        \cline{2-3} 
            &Answer Ver. &Health Fact Ver.  & Justification Ver. &Discriminative Task & Generative Task \\
    \hline
        \rowcolor{mygray} \multicolumn{6}{c}{Close-source Models} \\         
        
        GPT-4           & 66.15 & 78.60 & 16.28/16.27 & 86.63 & \textbf{27.09/23.71} \\ 
        Claude-3.5      & 11.54 & 62.04 & 14.77/16.45 & 82.14 & 17.80/20.02 \\  
        \midrule
        \rowcolor{mygray} \multicolumn{6}{c}{Open-source Models} \\  
        MEDITRON        & 25.23 & 32.66 & 11.58/15.78 & 60.83 & 4.42/14.08 \\  
        InternLM 2      & \textbf{99.23} & 76.94 & 8.75/14.69  & 84.67 & 15.84/19.01 \\ 
        Llama 3         & 94.77 & 63.89 & \textbf{16.52/16.49} & 63.85 & 21.31/22.75 \\ 
        Mistral         & 57.38 & 69.78 & 15.98/16.43 & 71.59 & 13.03/15.47  \\  
        \midrule
        MMedIns-Llama 3 & 97.08 & \textbf{79.55} & 12.71/14.65 & \textbf{86.71} & 23.52/25.17\\
    \bottomrule
    \end{tabular}
  \end{threeparttable}
  \vspace{4pt}
\end{table}


\subsubsection*{Natural Language Inference (NLI)}
Table~\ref{FactVerificationResults} shows the evaluation on medical Natural Language Inference (NLI) using the MedNLI textual entailment dataset. 
The results are measured with accuracy for the discriminative tasks~(selecting the right answer from a list of candidates) and BLEU/ROUGE metrics for the generative tasks~(generating free-form text answers).

InternLM 2 achieves the highest scores among the open-source LLMs, scoring 84.67. For the closed-source LLMs, GPT-4 and Claude-3.5 all show relatively high scores, with 86.63 and 82.14 accuracy scores respectively. 
Mistral shows the second best accuracy score of 71.59 among the open-source LLMs.
In the generative task, Llama 3 shows the highest consistency with the referenced ground truth with 21.31/22.75 for BLEU and ROUGE among the open-source models. 
Mistral and InternLM 2 exhibit mid-range performance in the generative task, with Mistral achieving BLEU/ROUGE scores of 13.03/15.47, InternLM 2 achieving 15.84/19.01 scores. Similarly, GPT-4 also performs well in the generative task format, resulting in 27.09/23.71 scores while Claude-3.5 is not ideal in this task.

\textbf{MMedIns-Llama 3} achieves the highest accuracy in the discriminative task, scoring 86.71, comparable with GPT-4. MMedIns-Llama 3 also excels in the generative task, with BLEU/ROUGE scores of 23.52/25.17, outperforming other models except the GPT-4.

\section{Discussion}
Overall, this paper makes several key contributions:

\subsection*{A Comprehensive Evaluation Benchmark – MedS-Bench} 

The development of medical LLMs has largely relied on benchmarks focused on multiple-choice question answering (MCQA). However, this narrow evaluation framework risks overlooking the broader capabilities required for LLMs in various clinical scenarios. In this work, we introduce \textbf{MedS-Bench}, a comprehensive benchmark designed to assess the performance of both closed-source and open-source LLMs across diverse clinical tasks, including those that require fact recall from the model or reasoning from given context. 
Our results reveal that while existing LLMs perform exceptionally well on MCQA benchmarks, they struggle to align with the actual clinical practice, particularly in tasks such as treatment planning and explanation. 
This finding underscores the need for further efforts to develop medical LLMs that are better suited to a wider range of clinical and medical scenarios beyond MCQA.

\subsection*{A New Comprehensive Instruction Tuning Dataset – MedS-Ins} 
We have developed \textbf{MedS-Ins}, a novel medical instruction tuning dataset, by extensively sourcing data from existing BioNLP datasets and converting these samples into a unified format, with semi-automated prompting strategies. 
Previous efforts have focused primarily on constructing question-answer pairs from daily conversations, exams, or academic papers, often neglecting the texts generated from real clinical practice. 
In contrast, MedS-Ins integrates a broader range of medical text sources, encompassing five primary text domains and 19 task categories, as illustrated in Figure~\ref{fig:topics_distribution}d. This systematic analysis on data composition is crucial for aligning LLMs with the diverse queries encountered in clinical practice.

\subsection*{A Strong Large Language Model for Medicine – MMedIns-Llama 3} 

On the model front, we demonstrate that by conducting instruction tuning on \textbf{MedS-Ins}, we can significantly enhance the alignment of open-source medical LLMs with clinical demands. Our final model, \textbf{MMedIns-Llama 3}, serves as a proof-of-concept, featuring a medium-scale architecture with 8 billion parameters, has exhibited a deep understanding of various clinical tasks and adapts flexibly to multiple medical scenarios through zero-shot or few-shot instruction prompts, without the need for further task-specific training. 
As evidenced by the results, our model outperforms existing LLMs, 
including GPT-4, Claude-3.5, across a range of medical benchmarks, 
covering different text sources.

\textbf{Limitations.} 
Here, we highlight the limitations of our paper and the potential improvements in future work.

\textit{First}, \textbf{MedS-Bench} currently covers only 11 clinical tasks, 
which does not fully encompass the complexity of all clinical scenarios. Additionally, while we evaluated six mainstream LLMs, some models remain absent from our analysis. To address these limitations, we plan to release an open leaderboard for medical LLMs alongside this paper. This initiative aims to encourage contributions from the community to continually expand and refine comprehensive benchmarks for medical LLMs. Specifically, this will involve updating the test set to better reflect real clinical demands and including a broader range of medical LLMs. By incorporating more task categories from diverse text sources into the evaluation process, we hope to gain a deeper understanding of the ongoing advancements in LLMs within the medical field.

\textit{Second}, although \textbf{MedS-Ins} now encompasses the widest range of medical tasks available, it remains incomplete, and certain practical medical scenarios may be missing. To address this, we have made all our collected data and resources available as open-source on GitHub. We encourage contributions from the broader AI4medicine community to help maintain and dynamically expand this instruction tuning dataset, similar to efforts for Super-NaturalInstructions in the general domain\cite{wang-etal-2022-super}. Detailed guidelines are provided on our GitHub page, and we will acknowledge every contributor involved in updating the dataset. The current limited number of tasks may explain why we have not yet observed the models exhibiting emergent abilities to generalize to unseen clinical tasks, a capability seen in LLMs trained on thousands of diverse tasks in the general domain~\cite{wang2022super,longpre2023flan}.

\textit{Third}, we plan to incorporate more languages into \textbf{MedS-Bench} and \textbf{MedS-Ins} to support the development of more robust multilingual LLMs for medicine. For now, although we include some multilingual tasks in both MedS-Bench and MedS-Ins, these resources are predominantly English-centered. 
Expanding to include a broader range of languages would be a promising future direction, ensuring that the latest advancements in healthcare AI can benefit a wider and more diverse range of regions equitably.

\textit{Finally}, all our code, data, and evaluation pipelines are open-sourced. We hope this work would inspire the medical LLM community to focus more on aligning these models with real-world clinical applications.

\clearpage
\section{Methods}

\subsection{Data Collection}
In this section, we describe the procedure for constructing \textbf{MedS-Ins}, as shown in Figure~\ref{fig:method}a. In order to organize the different tasks, we assign a domain tag and category tag to each task, the former denotes the domain covered by the instructions, while the category tag denotes the applicable task. We start by filtering the medical-related sentence in natural instruction datasets, followed by prompting specific BioNLP into free-text response formats.

\subsubsection*{Filtering Natural Instructions}
We start by filtering medical-related tasks from the 1,616 tasks collected in Super-NaturalInstructions~\cite{wang-etal-2022-super}. 
As this work focuses more on different natural language processing tasks in general-purpose domains, the granularity of classification is relatively coarse for the medical domain. 
We first extract all the instructions in ``Healthcare'' and ``Medicine'' categories, subsequently, we manually added more detailed granularity to the domain labels for them, while the task category remains unchanged. 

In addition, we found that many of the organized instruction fine-tuning datasets in the generic domain also cover some healthcare-related data, such as LIMA~\cite{zhou2024lima} and ShareGPT~\cite{li2023compressing}. To filter out the medical part of these data, we used InsTag~\cite{lu2023instag} to classify the domain of each instruction at a coarse granularity. Specifically, InsTag is an LLM, specialized for tagging different instruction samples. Given an instruction query, it will analyze which domain and task it belongs to. 
Finally, by filtering instruction datasets in the general domain, we collect 37 tasks, for a total of 75373 samples.

\subsubsection*{Prompting Existing BioNLP Datasets}
In the literature, there exists many excellent datasets on text analysis in clinical scenarios. However, as most datasets are collected for different purposes, like classification or text completion, they can not be directly used for training large language models. Here, we convert these existing former medical NLP tasks into a format that can be used for training generative models, naturally adding them into instruction tuning. 

Specifically, we use MIMIC-IV-Note as an example, 
which provides high-quality structured reports with both findings and impressions, they are used  as a proxy task for text summarization, 
where impressions act as an abstract summary of the findings. 
We first manually write prompts to define the task, for example,   \textit{``Given the detailed finding of Ultrasound imaging diagnostics, summarize the note's conclusion in a few words.''}. 
Considering the diversity for instruction tuning, 
we ask 5 individuals to independently describe a certain task with 3 different prompts. This results in 15 free-text prompts for each task, with similar semantic meanings but as varied as possible in wording and format. 
Then, inspired  by the Self-Instruct~\cite{wang2022self}, we use these manually written instructions as seed prompts and asked GPT-4~\cite{jpn-med-exam_gpt4} to rewrite more task instructions based on the following prompt:
\begin{mdframed}[backgroundcolor=gray!20]
Rewrite the following instruction definition directly. You can change the wording, but keep the meaning the same. Output the rewritten definition directly without any additional information.
\end{mdframed}

\begin{figure}[t]
    \centering
    \includegraphics[width=\linewidth]{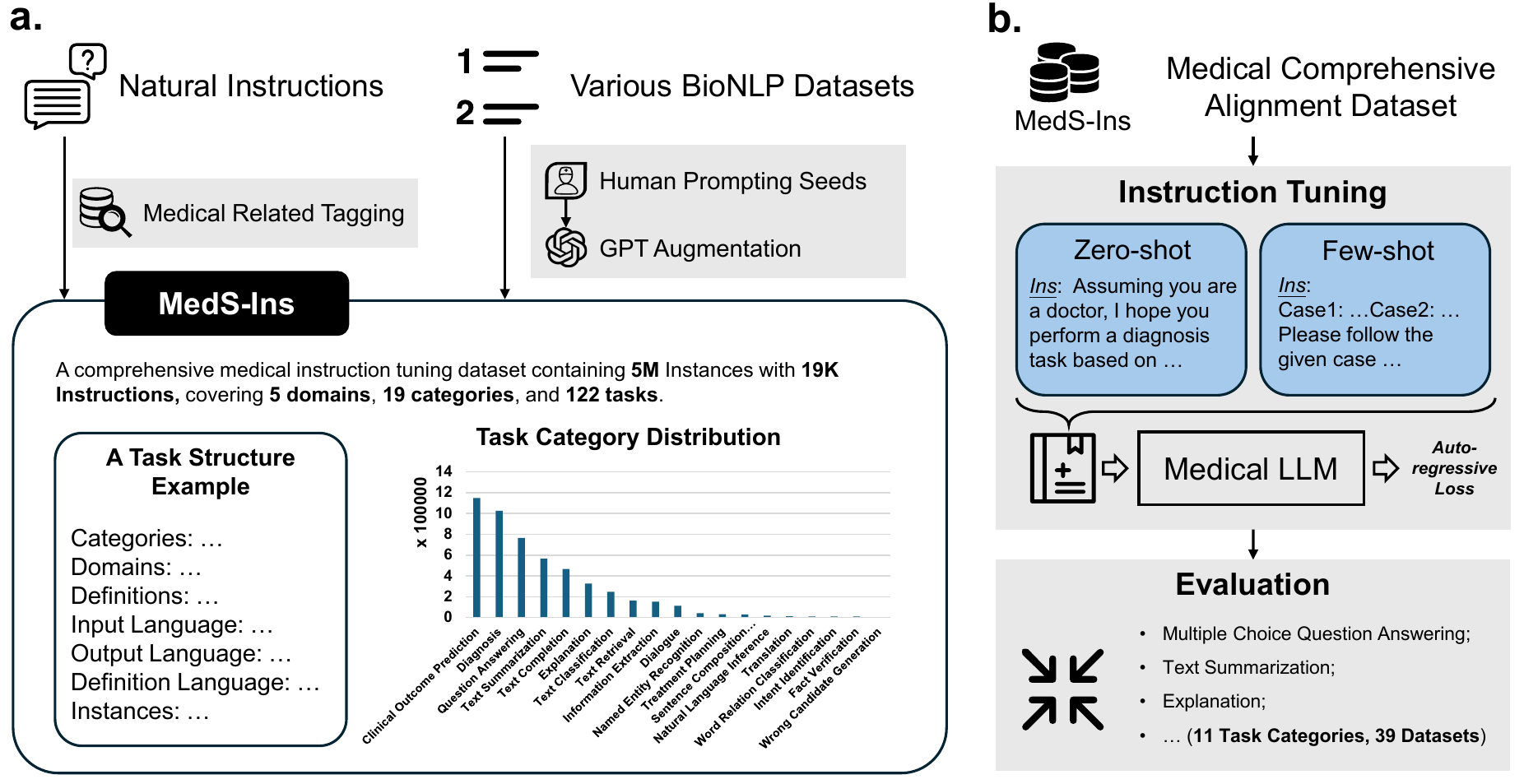}
    \caption{\textbf{The pipeline of our method.} \textbf{a} The data collection pipeline. We mainly collect data through filtering natural instructions and prompting well-organized BioNLP datasets. \textbf{b} The training and evaluation pipeline for our model leveraging the collected MedS-Ins. We leverage the instruction tuning training method to combine different datasets and evaluate the final model on multiple benchmarks comprehensively.}
    \label{fig:method}
\end{figure}

Finally, for each task, we will describe it with 7 key elements as shown at the bottom of Figure~\ref{fig:method}a, \emph{i.e.}, \textit{\{``Categories'', ``Domains'' ``Definitions'', ``Input Language'',  ``Output Language'', ``Instruction Language'' and ``Instances''\}}, 
where ``Definition'' consists of the manually written or GPT-4 enhanced instruction to describe the tasks, ``Input Language'', ``Output Language'', and ``Instruction Language'' respectively describe the languages, such as English or Chinese, used in the corresponding components of a specific instance of this task. ``Categories'' and ``Domains'' describe what text domains and categories the task belongs to. Finally, in ``Instances'', different training or evaluation instances with Input and Output contents are stored.

Through the above procedure, we prompt an extra 85 tasks into a unified free-form question-answering format, combined with the filtered data, resulting in a totaling 5M instances with 19K instructions, covering 122 tasks, termed as \textbf{MedS-Ins}~(the detailed 122 task information can be found in Supplementary Sec.~\ref{sec:detail MedSIns}), which has shown to significantly improve the LLMs on clinical tasks. 



\subsection{Model Training}
In this section, we detail the training procedure, as shown in Figure~\ref{fig:method}b. We take the same approach as our previous work~\cite{wu2024pmc, qiu2024towards}, 
which have shown that further auto-regressive training on medical-elated corpus can inject medical knowledge into the models, thus allowing them to perform better in different downstream tasks. 
We start from a multilingual LLMs base model~(MMed-Llama 3~\cite{qiu2024towards}), and further train it with comprehensive instructions from \textbf{MedS-Ins}.



\subsubsection*{Instruction Tuning} 
Given the base model, trained on a large-scale medical corpus with auto-regressive prediction, we further fine-tune it to better follow human instructions or prompts. Considering an input sequence with an instruction \( I \) and a context \( C \), and an output sequence \( O \), the model is trained to maximize the probability:
\begin{equation}
    P(O | C, I) = \prod_{t=1}^{|O|} P(o_t | o_1, o_2, ..., o_{t-1}, C, I; \theta)
\end{equation}
Similarly, the loss function used in instruction tuning is cross-entropy loss and can be calculated as follows:
\begin{equation}
    \text{Loss} = -\sum_{t=1}^{|O|} \log P(o_t | o_1, o_2, ..., o_{t-1}, C, I; \theta)
\end{equation}
The key insight here is to construct diverse instructions, 
that enables the model to robustly output the preferred answers. 
Here, we mainly consider two types of instructions, 
namely, zero-shot and few-shot prompting.

\noindent \textbf{Zero-shot Prompting.} 
Here, the $I$ contains some semantic task descriptions as hints, and the model is therefore asked to directly answer the questions based on its internal model knowledge. In our collected \textbf{MedS-Ins}, the ``Definition'' contents for each task can be naturally used as the zero-shot instruction input. Due to the coverage of a wide range of different medical task definitions, the model is expected to learn the semantic understanding of various task descriptions. The input template is as follows:
\begin{mdframed}[backgroundcolor=gray!20]
\{INSTRUCTION\} \\
Input: \{INPUT\}
\end{mdframed}

\noindent \textbf{Few-shot Prompting.} 
Here, the $I$ contains the few-shot examples, that allow the model to learn the input-output mapping on the fly. We simply obtain such instructions by randomly sampling other cases from the training set of the same task, and organizing them using a straightforward template as follows: 
\begin{mdframed}[backgroundcolor=gray!20]
Case1: Input: \{CASE1\_INPUT\}, Output: \{CASE1\_OUTPUT\}\\
\dots \\
CaseN: Input: \{CASEN\_INPUT\}, Output: \{CASEN\_OUTPUT\}\\
\{INSTRUCTION\} \\
Please learn from the few-shot cases to see what content you have to output.\\
Input: \{INPUT\}
\end{mdframed}

\subsubsection*{Implementation Details}
We conduct all our experiments using \texttt{PyTorch} framework and \texttt{Transformers} python package. Specifically, we set the maximum length to 2048, and pad the sequence to the longest case with padding tokens in a batch. We employ the Fully Sharded Data Parallel~(FSDP) implemented with \texttt{Transformers.trainer} function to save the memory cost per GPU. We also adopt BF16 as default training precision and gradient checkpointing~\cite{Chen2016TrainingDN} techniques to optimize memory usage.  We use a global batch size of 128 and a learning rate of 1e-5. We choose the medical-knowledge-enhanced model MMed-Llama 3 in our previous work as the foundation model. We further train the model by supervised fine-tuning on Meds-Ins for 5 Epoch with 32 Ascend910B for 383.5 hours.



\subsection{Baselines}
Here, we provide details for the baseline large language models~(LLMs). Note that, we evaluate all models in few-shot settings, as we observe that open-source models struggle to complete zero-shot evaluation.
Specifically, \textbf{three} example cases are given to the model, the detailed prompting strategy can be found in Supplementary Sec.~\ref{Sec:Evaluation Settings}.

The first category includes the powerful closed-source LLMs, known for their robust performance in the general domain. We evaluate these models on various medical-specific tasks:

\vspace{-0.2cm}
\begin{itemize}
    \setlength\itemsep{0.3em}
    \item \textbf{GPT-4}~\cite{jpn-med-exam_gpt4}, developed by OpenAI, stands for one of the most sophisticated LLMs to date. 
    It is renowned for its strong capabilities in language processing in general domains, including medical applications.
    
    \item \textbf{Claude-3.5}~\cite{claude}, developed by Anthropic, is a frontier AI language model designed to be secure, trustworthy, and reliable. It exhibits advanced reasoning capabilities that enable it to perform complex cognitive tasks effectively. We adopt the Claude-3.5-Sonnet for comparison, which is claimed as the best model among the Claude family.
\end{itemize}

The second category comprises the mainstream open-source LLMs:
\vspace{-0.2cm}
\begin{itemize}
    \setlength\itemsep{0.3em}
    \item \textbf{Llama\ 3}~\cite{touvron2023llama}, developed by Meta AI, is one of the most notable open-source LLMs globally. 
    As part of the LLaMA series, it is designed for high performance in natural language processing tasks, with enhancements over its predecessors in accuracy and contextual understanding. 
    Since our trained model MMedIns-Llama 3 is based on the 8B version, for a fair comparison, we also mainly consider the 8B version for Llama 3.
    
    \item \textbf{Mistral}~\cite{jiang2023mistral}, developed by Mistral AI, is an innovative open-source LLM that claims superiority over Llama 2 13B across all evaluated benchmarks. Similarly, for a fair comparison, we also consider its 7B version. 
    
    \item \textbf{Internlm\ 2}~\cite{cai2024internlm2}, developed by Shanghai AI Laboratory, stands out as a leading open-source multilingual LLM, showcasing exceptional performance, particularly in English and Chinese. We adopt the 7B version of Internlm\ 2. 
    
\end{itemize}

The third category of models we choose is the open-sourced medical LLMs, which have been further trained on medical data.

\vspace{-0.2cm}
\begin{itemize}
    \item \textbf{MEDITRON}~\cite{chen2023meditron} is a large-scale medical LLM with 7B parameters, further pre-trained on Llama 2. 
    It leverages 21.1M medical papers, guidelines for further pre-training, and supervised finetuning on different MCQA datasets with context and chain-of-thought prompt styles. 
\end{itemize}

\subsection{Metrics}

In this section, we delineate the metrics employed across various tasks and categories within our benchmark. 

\noindent \textbf{Accuracy}: For tasks requiring the model to select a single correct answer from multiple choices, we employ `accuracy' as a direct metric. This metric is applied to tasks \textit{MedQA, MedMCQA, and MMedBench} in Multilingual Multiple-choice Question-answering; \textit{participant, intervention, and outcome extraction in PICO, drug dose extraction in ADE, and patient information extraction in PMC-patient} for Information Extraction. 
It is also used in \textit{SEER} for Treatment Planning, \textit{DDXPlus} for Diagnosis, \textit{MIMIC4ED} for Clinical Outcome Prediction, \textit{PubMedQA and PUBLICHEALTH Verification} for Fact Verification, as well as \textit{MedNLI textual entailment discriminative tasks} for NLI.

\noindent \textbf{Precision, Recall, F1 Score}: For tasks where the model is required to select multiple correct answers, we utilize Precision, Recall, and the F1 Score. These metrics are relevant for\textit{BC4Chem and BC5Chem for chemical recognition, BC5Disease for disease recognition, Species800 for organism recognition} in Named Entity Recognition (NER), and \textit{HoC} in Classification.

\noindent \textbf{BLEU, ROUGE}: For tasks necessitating the generation of free-form text, which are inherently more challenging to evaluate, we utilize BLEU and ROUGE metrics to assess the similarity between the generated text and the ground truth. Specifically, we use BLEU-1 and ROUGE-1 by default if no other statements in this paper. These tasks include \textit{MedQSum, RCT-Text, MIMIC-CXR, MIMIC-IV} for Text Summarization; \textit{EBMS} for Fact Verification; \textit{PUBLICHEALTH Explanation, Do, BioLORD and MMedBench}  for Concept Explanation / Rationale; along with \textit{generative tasks in textual entailment in MedNLI} for NLI.


\section{Conclusion}
In this paper, we construct a new medical evaluation benchmark, \textbf{MedS-Bench}, which contains 11 clinical tasks beyond multiple choice problems, and thus aims to provide a more comprehensive assessment for clinical scenarios. We evaluate 6 mainstream LLMs, showing that existing models are fragile in complex real medical usage. Moreover, to alleviate the gap, we systematically collect a wide range of existing biomedical NLP datasets, across 5 text domains and with 122 tasks, resulting in \textbf{MedS-Ins}. We conduct further instruction tuning on the multilingual base model from our previous work, resulting in a new medical LLM, namely \textbf{MMedIns-Llama 3}. Our final model demonstrates superior performance across almost all tasks on the benchmark. 
To further advocate the development of LLMs towards clinical tasks,
we have fully opened our instruction tuning set and encourage more researchers to participate in building up more comprehensive medical instruction data together.
Moreover, we have built up a leaderboard for \textbf{MedS-Bench}, that will dynamically update and expand the test set to monitor progress, for better adapting general LLMs towards the medical domain. 


\clearpage
\bibliographystyle{sn-mathphys} 
\bibliography{references} 
\clearpage
\appendix

\section{Supplementary}
\subsection{Evaluation Settings}
\label{Sec:Evaluation Settings}
To assess the model performance on different tasks, there are mainly two prompting strategies, \emph{i.e.}, zero-shot prompting and few-shot prompting. In our experience, we find that in a zero-shot setting, though the detailed descriptions of different datasets are given, 
existing LLMs can hardly understand the strict format or clinical requirements, even for the strongest close-source model GPT-4 and Claude-3.5. Thus, all evaluations are based on a few-shot~(3-shot) prompting strategy following the below template:
\begin{mdframed}[backgroundcolor=gray!20]
Case1: \dots; \\
Case2: \dots; \\
Case3: \dots; \\
\dots \\
\{Manually Written Definitions\} \\
Please learn from the few-shot cases to see what content you have to output. \\
\{Input Case\}
\end{mdframed}

\subsection{Task Category Details}
\label{Sec: Benchmark Details}
In this paper, we conclude 19 medical-related task categories in total. For evaluation, we adopt 11 categories, covering 28 datasets, as shown by Supplementary Table~\ref{tab:benchmark_statistics}. 
To ensure the integrity of our evaluation, we deliberately exclude the test set from the source dataset during the prompting of instances. This test set is subsequently utilized as a benchmark for assessing the performance of LLMs. For datasets that do not have a predefined train-test split, we randomly partition the dataset into training and testing subsets at an 9:1 ratio. 
Notably, as the evaluation involves commercial models, for example, GPT-4 and Claude 3.5, it is extremely costly to adopt the original large-scale test split. Therefore, for some benchmarks~(as will be detailed in the following), we randomly sampling a number of test cases. For the MCQA tasks, we use all of them, as the close-source models have been evaluated by other papers.

In the following, we will introduce each task category in detail, and further list out the used benchmark names, if the category has been considered in the evaluation.

\noindent \textbf{Medical Multi-choice Question Answering~(MCQA).} 
Medical Multiple-choice Question Answering is a straightforward method for evaluating the performance of medical models. During evaluation, the model is required to select the correct answers. This method has been widely adopted as an evaluation benchmark to assess whether the models contain enough medical knowledge. Therefore, we also adopt the popular multiple-choice question-answering benchmarks, covering both English and multilingual medical domains.

\vspace{-8pt}
\begin{itemize}
    \setlength\itemsep{0.5em}
    \item \textbf{MedQA~\cite{jin2021disease}} is a collection of medical multiple-choice questions, providing both 4-option and 5-option versions. It covers three languages, \emph{i.e.}, English, Simplified Chinese, and Traditional Chinese. In this task, to be aligned with existing works~\cite{wu2024pmc,singhal2023large}, 
    we adopt the 4-option \textbf{English} version with official split. There are 1273 samples in the testset.
    
    \item \textbf{PubMedQA~\cite{jin-etal-2019-pubmedqa}} is an \textbf{English} question-answering medical dataset based on PubMed abstracts. The task of PubMedQA is to answer research questions with yes/no/maybe, which can also be treated as a close-domain multiple-choice question-answering problem. Officially, it is split into three subsets: 1K manually labeled pairs (PQA-L), 61.2K unlabeled pairs (PQA-U), and 211.3K artificially generated pairs (PQA-A). Following existing works~\cite{diao2023lmflow}, we adopt PQA-L as the testing set. There are 1000 samples in the testset.
    
    \item \textbf{MedMCQA~\cite{pmlr-v174-pal22a}} is a large-scale {English} multiple-choice question-answering samples. MedMCQA has more than 194k high-quality AIIMS \& NEET PG entrance exam multiple-choice questions covering 2.4k healthcare topics and 21 medical subjects are collected with an average token length of 12.77 and high topical diversity. The official train split contains 182,822 questions, and the test split contains 4183 questions. Each question has 4 choices. We adopt the official test split to evaluate our model.
    \item \textbf{MMedBench~\cite{qiu2024towards}} is a {multilingual} medical multiple-choice question-answering datasets covering 6 main languages, \emph{i.e.}, English, Chinese, French, Japanese, Russian and Spanish. It collects multiple-choice questions from the following datasets: \textbf{MedQA}~\cite{jin2021disease}, \textbf{lgakuQA}~\cite{jpn-med-exam_gpt4}, \textbf{FrenchMedMCQA}~\cite{labrak:hal-03824241}, \textbf{RuMedDaNet}~\cite{blinov2022rumedbench},\textbf{Head-QA}~\cite{vilares-gomez-rodriguez-2019-head}. We have to emphasize that its English evaluation part is the same as MedQA, thus the reporting scores on it may be overlapped with MedQA for English. We followed its official split to test our model. That is, 3426 samples for Chinese, 199 samples for Japanese, 622 samples for French, 256 samples for Russian, and 2742 samples for Spanish.
\end{itemize}

All the above benchmarks are based on multiple-choice questions. Therefore, the model ability can be easily assessed via the  default Accuracy score as previous work~\cite{qiu2024towards,wu2024pmc,diao2023lmflow,singhal2023large,singhal2023towards}.

\noindent \textbf{Text Summarization.} 
Text summarization denotes a generative task, 
where a model is tasked to extract key information from extensive pieces of text. This task is particularly prevalent in the clinical field, for example, quickly extracting essential information from lengthy examination reports or helping clinicians and patients better understand complex medical reports. We evaluate the model's performance using the following benchmarks. Notably, different from other tasks, text summarization commonly involves longer contexts which is more costly for evaluating the close-source models, thus we adopt smaller sampling cases~(100) for evaluation. 

\vspace{-8pt}
\begin{itemize}
    \setlength\itemsep{0.5em}
    \item \textbf{MedQSum~\cite{lee2021mnlp}} is a large, freely available database of de-identified health-related data associated with over fifty thousand patients. It contains notes, reports, and electronic health records (EHR) of the patients, formatted with a fixed template. For instance, imaging diagnostic reports always start with detailed findings and end with a conclusion. We reconstruct this dataset by using the detailed findings as input to the model and expect the model to generate a summary, with the conclusion of the note serving as the ground truth. During testing, we randomly sample 100 instances for evaluation.
    
    \item \textbf{RCT-Text~\cite{RCTText}} is a dataset designed to summarize medical evidence from various clinical studies as presented in literature reviews. For this task, we provide the LLM with the titles and abstracts of studies and task the model to output the primary conclusions of each study. During testing, we randomly sample 100 instances for evaluation.
        
    \item \textbf{MIMIC-CXR~\cite{MIMICCXR}} is a comprehensive, publicly available dataset that includes chest radiographs, each paired with structured labels derived from free-text radiology reports. The dataset encompasses 227,827 radiology reports, formatted with one template. For this task, we provide the LLM with descriptions highlighting key aspects of the chest X-ray images. The model is then required to summarize the pertinent findings from these descriptions. During testing, we randomly sample 100 instances for evaluation.
    
    \item \textbf{MIMIC-IV Report~\cite{johnson2023mimic}} is a sub-set of MIMIC-IV. Specifically, MIMIC-IV is a large deidentified medical dataset of patients admitted to the emergency department or an intensive care unit at the Beth Israel Deaconess Medical Center in Boston, MA. It contains data for more than 65K patients admitted to an ICU and over 200K patients admitted to the emergency department. In this large-scale medical informatics database, there is a radiology note split that contains 2,321,355 deidentified radiology reports for 237,427 patients. For each radiology report, inspired by existing works~\cite{van2024adapted,zhang2018learning}, 
    we treat the impression part as a general summarization of the findings. We randomly sampled 100 cases for each covered modality and body region part, like Chest CT, Brain MRI, etc., resulting in 1430 cases~(for some parts, the test split has fewer than 100 cases, we directly keep all.) for evaluation, covering different anatomies and modalities.
\end{itemize}

In all the above benchmarks, the model is required to generate free-form texts. We evaluate the quality of the generated text by calculating the similarity between the generated text and the ground truth with BLEU~\cite{papineni2002bleu} and ROUGE~\cite{lin2002manual}.

\noindent \textbf{Information Extraction.} 
In contast to summarization that aims at concluding the main information from given contexts, information extraction instead is expected to extract the detailed clinical or medical information from long texts based on various user queries.

\vspace{-8pt}
\begin{itemize}
    \setlength\itemsep{0.5em}
    \item \textbf{ADE Corpus~\cite{gurulingappa2012development}} provides information on the drug and its corresponding adequate dose within each sentence. For the drug dose extraction task, we input into the model both the sentence and the drug name mentioned therein. The model is then required to identify the dosage levels of the specified drug. 
    In this benchmark, we directly use the dataset prompted by Super-Instruction~\cite{wang2022super} with a few cases, and divide it at a ratio of 9:1 for instruction tuning and evaluation. During testing, we evaluate using the entire test set, which consists of 23 instances.
    
    \item \textbf{PICO~\cite{nye-etal-2018-corpus}} consists of 5,000 abstracts from medical articles on the randomized controlled clinical trials, annotated to identify text spans that describe the patient population enrolled, the interventions studied and their comparisons, and the outcomes measured. We employ this dataset to develop three tasks. The first task, Outcome Extraction, requires the model to identify phrases in a study report sentence that provide information about the study's outcomes. The second task, Intervention Extraction, involves the model identifying phrases in a study report sentence that describe the study’s interventions. The final task, Participant Extraction, demands the model to extract phrases from a study report sentence that detail information about the study's participants. 
    In this benchmark, we directly use the dataset prompted by Super-Instruction~\cite{wang2022super} with a few cases, and divide it at a ratio of 9:1 for instruction tuning and evaluation. During testing, we evaluate using the entire test set, which consists of 43 instances for each task.

    \item \textbf{PMC-patient basic information extraction~\cite{PMCPaitent}} is a pioneering collection that includes 167,000 patient summaries, extracted from case reports in PubMed Central (PMC). This dataset has been further enriched with annotations detailing basic patient information. For this specific task, we provide the Large Language Model (LLM) with an extensive clinical text about the patient, and require the model to accurately extract information regarding the patient's gender and age. During test, we randomly sample 1500 instances for evaluation.
\end{itemize}

\noindent \textbf{Explanation.} Explanation considers the task of providing a detailed justification or description of a clinical claim or concept. It is useful in clinical to help physicians or patients better understand professional medical terminologies or claims from other specialists. Our benchmark includes the following datasets.

\vspace{-8pt}
\begin{itemize}
    \setlength\itemsep{0.5em}
    \item \textbf{PUBHEALTH Explanation~\cite{kotonya2020explainable}} is a comprehensive dataset geared towards explainable automated fact-checking of public health claims. The task requires models to furnish an explanation for a specified claim using supporting material from the provided paragraph. During testing, we use the entire test set consisting of 1235 samples.
    \item \textbf{HumanDiseaseOntology (DO)~\cite{schriml2019human}} is a database providing the biomedical community with consistent, reusable, and sustainable descriptions of human disease terms, phenotype characteristics, and related medical vocabularies. In this task, we ask the model to explain the specified given medical professional entity, and the description from the database functions as the ground truth. During testing, we use the entire test set of the dataset, which consists of 1115 instances.
    \item \textbf{BioLORD Explanation~\cite{remy2022biolord}} comprises pairs of biomedical concepts names and descriptions. In this task, we challenge models to elaborate on concise concepts by generating long, detailed definitions. During testing,  we use the entire test set of the dataset, which consists of 1115 instances.
    \item \textbf{MMedBench-Rationale~\cite{qiu2024towards}} is a multilingual medical multiple-choice question and answering benchmark along with rationale for each sample. In this task, we will give the LLMs a question and its answer and require them to generate an explanation statement or step-by-step reasoning. 
    During testing, We utilize the entire test set of the dataset, comprising detailed allocations of samples for each language: 200 for Chinese, 195 for English, 189 for French, 190 for Japanese, 173 for Russian, and 189 for Spanish.
\end{itemize}

For these tasks, the models are required to generate free-form text. The quality of the generated text is evaluated by comparing it to the ground truth using BLEU and ROUGE metrics.

\noindent \textbf{Named Entity Recognition.}
Named Entity Recognition (NER) in the biomedical field is a specialized task in natural language processing that identifies and classifies entities in biomedical texts, such as research articles, clinical notes, and patient records. These entities include gene and protein names, diseases, chemicals, drugs, anatomical structures, and other medical terms. This task can assist clinicians to structure clinical free-form texts, thus enables more readable documents. 
Due to the required domain-specific knowledge, 
this task poses significant challenges, particularly for large language models. Our benchmark evaluates performance across the following benchmarks.

\vspace{-8pt}
\begin{itemize}
    \setlength\itemsep{0.5em}
    \item \textbf{BC4Chem~\cite{savery2020chemical}} is a dataset comprising 10,000 PubMed abstracts with a total of 84,355 chemical entity mentions, manually annotated by expert chemistry literature curators. In this task, the LLM is to input an abstract, 
    and is required to recognize the name of the chemical. 
    During testing, we randomly sample 1500 instances for evaluation.
    
    \item \textbf{BC5CDR~(BC5Chem and BC5Disease)~\cite{li2016biocreative}} is a widely-used resource in biomedical natural language processing, annotated for chemical and disease entities and their relationships. This benchmark is split into 2 task according to the category of the entity. In the first category, the LLM is required to recognize the name of the chemical. In the other category, the LLM is required to write the name of the disease. During testing, we randomly sample 1500 instances for each task.
    
    \item \textbf{Species800~\cite{pafilis2013species}} comprises 800 PubMed abstracts in which organism mentions were identified. In this task, the LLM is input an abstract and is required to recognize the name of the organism. During testing, we randomly sample 1500 instances for evaluation.
\end{itemize}

Based on the entity numbers, we categorize these datasets into two cases, \emph{i.e.}, datasets containing exactly one entity per sentence, and those that may contain arbitrary entities~(multiple or none), challenging the model to differentiate the text structure effectively. For the former cases, performance is assessed using accuracy metrics, and for the latter ones, 
Recall, Precision, and mixed F1 scores are used.

\noindent \textbf{Diagnosis.} Diagnosis is an important and basic task for clinical practices. The context of patients will be given in this task, with EHR or free-text format, and models are expected to predict the final diseases from a suspicious disease list or directly give positive or negative judgment for a certain targeting disease proposed by the user. The following benchmark is used for evaluation:
\begin{itemize}
    \item \textbf{DDXPlus~\cite{fansi2022ddxplus}} is a pioneering large-scale dataset tailored for Automatic Symptom Detection (ASD) and Automatic Diagnosis (AD) systems in the medical domain, featuring synthesized patient data from a proprietary medical knowledge base and commercial AD system. In this task, the model must make diagnostic decisions based on dialogues, selecting from a provided list of potential diagnoses. During testing, we randomly sample 1500 instances for evaluation.
\end{itemize}
We report the accuracy scores to reflect the models' diagnosis ability.

\noindent \textbf{Treatment Planning.} Similar to diagnosis, treatment planning is also a common clinical routine. The input texts for this task are also the context of a patient, while the output of models is expected to be treatment recommendations, like medication or therapy types. Treatment is a complex task in clinical, as the detailed treatment plan can be highly customised and detailed in free-form texts, as an evaluation sub-task, we simplify this task into picking the best treatment decision from a given close-set. 
The following benchmarks are considered:
\begin{itemize}
    \item \textbf{SEER~\cite{dubey2023using}} is a treatment planning dataset leveraging the  Surveillance, Epidemiology, and End Results~(SEER) custom breast cancer databases.  It collects 32,922 patient records and for each record with 19 attributes where the recommended treatment plans are noted, commonly in five types, \emph{i.e.}, Surgery Chemotherapy, Hormonal therapy, Biological therapy, and Radiation therapy. During testing, we randomly sample 1500 instances for evaluation.
\end{itemize}
As we have simplified this task into a close-set recommendation format, we can therefore assess the model's performance with accuracy scores.

\noindent \textbf{Clinical Outcome Prediction.} 
Beyond diagnosis and treatment,
there are also other tasks to help clinicians make clinical decisions, for example, disease risks and survival outcomes. This task judges whether the models can provide assistance in accurate clinical decision-making. The following benchmark is considered in our paper.

\vspace{-8pt}
\begin{itemize}
    \item \textbf{MIMIC4ED Benchmark~\cite{xie2022benchmarking}} provides a standardized benchmark derived from the Medical Information Mart for Intensive Care IV-Emergency Department (MIMIC-IV-ED) database, including data covering several key prediction tasks within the emergency department. We use this dataset for the comparison of LLMs for predicting clinical outcomes in emergency medicine. According to the differential of the categories of the clinical output, we split it into 3 tasks. We give the model an EHR of a patient and require the model to predict whether the patient needs, or the patient may revisit the emergency department in 72 hours, or the patient should be classified into a critical triage queue, separately. During testing, we randomly sample 1500 instances for each task.
\end{itemize}
Considering that, in this paper, the most clinical output is binary~(yes or no), we can assess the final model performance with commonly binary classification metrics, like accuracy, precision, recall, and F1 scores.

\noindent \textbf{Text Classification.}
Text classification aims to classifying a given text into a certain category. This is useful for clinical text collation and identifying public attitudes towards certain clinical processes.  The following datasets are used here.
\begin{itemize}
    \item \textbf{HoC~\cite{baker2016automatic}} is a specialized dataset containing 1,852 PubMed publication abstracts, which are expertly annotated according to a taxonomy. In this task, LLM is required for you to classify the hallmarks of cancer according to the given biomedical publication abstracts. During testing, we use the entire test set of the dataset, which consists of 158 instances for each task.
\end{itemize}

The above benchmark typically involve multi-class classification, thus, we use the accuracy as evaluation metric.

\noindent \textbf{Fact Verification.} Fact verification involves detecting some basic knowledge errors in a given context. 
This plays a role for large-scale verification on biomedical documents.
\begin{itemize}
    \setlength\itemsep{0.5em}
    \item \textbf{PUBHEALTH Verification~\cite{kotonya2020explainable}} is a dataset designed for explainable automated fact-checking of public health claims. Beyond asking models to explain the fact details, it can also be modified as a fact verication task, \emph{i.e.}, given a paragraph along with a related claim and models are required to determine whether the claim contradicts the evidence provided in the paragraph. During testing, we use the entire test set of the dataset, which consists of 650 instances for each task.

    \item \textbf{PubMedQA Verication~\cite{jin-etal-2019-pubmedqa}} is a question-answering medical dataset based on PubMed abstracts. Unlike direct use in a Question Answering benchmark, we provide the LLM with a passage accompanied by a question and its answer. The Task for LLM is to verify if the answer answers the question. During testing, we use the entire test set of the dataset, which consists of 1188 instances for each task.

    \item \textbf{EBMS justification verification~\cite{molla2011development}} During testing, we use the entire test set of the dataset, which consists of 304 instances for each task.
\end{itemize}





\noindent \textbf{Natural Language Inference~(NLI).}  Natural Language Inference (NLI) is one of the critical tasks for understanding natural language. The objective of NLI is to determine if a given hypothesis can be inferred from a given premise. In medical this task is useful for making inferences from the textual medical history of patients.

\begin{itemize}
    \item \textbf{MedNLI~\cite{romanov2018lessons}} 
    is a nature language inference dataset for clinical domain, which is grounded in patient medical histories and annotated by medical professionals. 
    We have divided the dataset into two tasks: `Entailment~(discriminative)' and `Entailment~(generative)'. 
    In the discriminative task, the model is initially presented with a formal clinical premise, such as condition descriptions or quantitative results. It is then given a hypothesis statement and must determine whether the hypothesis can be logically inferred from the clinical condition. In the generative task, the model is presented with a formal clinical premise and is required to generate a coherent and contextually correct hypothesis statement based on the given premise. During testing, we use the entire test set of the dataset, which consists of 1422 instances for each task.
\end{itemize}

\noindent \textbf{Text Retrieval.} Text retrieval refers to extracting the referenced text from a long given text, which is similar as introduced in Sec.~\ref{categories}. It can be used to match some similar cases from a large-scale case basis or related medical knowledge from encyclopedias for a given textual query. 

\noindent \textbf{Intent Identification} involves understanding and recognizing the intent behind user inputs. In clinical scenarios, patients may propose various queries, and the basic step for analysis is to distinguish their general intent like `treatment' or `disease'.  This can help the doctor understand the patient's request quickly and accurately, speeding up the process. 

\noindent \textbf{Word Relation Classification} involves classifying whether two given medical words are linked with some specific relation, like what drugs may interact with others or what symptoms may link with some diseases. This task can inject the basic medical knowledge about professional terminologies into models.

\noindent \textbf{Translation} relates to translating text from one language to another. It can help bridge language barriers between doctors and patients, as well as translate the latest medical research and literature into different languages, facilitating the global sharing and dissemination of medical knowledge.

\noindent \textbf{Dialogue} involves constructing systems capable of engaging in natural conversations with humans. Clinically, this can aid in the initial collection of patient histories and descriptions of symptoms, improving the efficiency of consultations. It holds significant importance for achieving personalized and precise medical care.

\noindent \textbf{Sentence Composition Analysis} involves decomposing a medical sentence into a structural linking graph, a task that can enhance LLMs' understanding of complex medical sentences. By doing so, it improves the model's ability to comprehend and process medical-related sentences more effectively.

\noindent \textbf{Wrong Candidate Generation} involves generating false answers to user questions, which, while not useful in real clinical contexts, serves as a fundamental ability for LLMs. By leveraging this capability, LLMs can identify outputs that are completely unacceptable and can also automatically generate data for other task categories, such as fact verification and multiple-choice question answering (MCQA).

\noindent \textbf{Text Completion} relates to automatically finishing some clinical texts. In healthcare scenarios, staff are asked to fill a large number of records or reports in daily workflow. Leveraging LLMs to aquatically initial a reference can improve the efficiency critically. This is a more general task category compared to others, we mainly include the tasks falling out the above introduced categories into this type.

\subsubsection{The Detailed Statistic of Our Benchmark}

In Table~\ref{tab:benchmark_statistics}, we further demonstrate the detailed token length of different task category in our combined benchmark, along with its corresponding language model abilities. 

\begin{sidewaystable}[!htb]
\centering
\begin{threeparttable}[b]
\footnotesize
\tabcolsep=0.10cm
   \vspace{10pt}
    \caption{The statistic details of used benchmarks.}
    \label{tab:benchmark_statistics}
\begin{tabular}{l|p{5cm}|l|p{1.2cm}|l|l|p{1.2cm}|p{1.2cm}}
\toprule
\multirow{2}{*}{Task Categories} & \multirow{2}{*}{Benchmarks}                        & \multirow{2}{*}{Assessed Ability} & \multicolumn{4}{c|}{Averaged Token Length}                                                                         & \multirow{2}{*}{Metrics} \\ \cline{4-7}
                                 &                                                    &                                   & Input                   & \multicolumn{1}{c}{Output} & \multicolumn{1}{c}{Definition} & \multicolumn{1}{c|}{Total} &                          \\ \midrule
Multi-choice Question Answering  & MedQA, PubMedQA, MedMCQA, IgakuQA, RuMedBench, FrenchMedMCQA, HeadQA & Recalling facts from \textbf{Pre-training} & 144 & 15                         & 21                             & 180                       & Accuracy    \\ \midrule
Explanation/Rationale  & PUBHEALTH Explanation, Human Disease Ontology (DO), BioLORD Explanation, MMedBench-Rationale & Recalling facts from \textbf{Pre-training} & 243 & 50                        & 46                             & 329                       & BLEU, ROGUE \\ \midrule
Diagnosis  & DDXPlus  & Recalling facts from \textbf{Pre-training} & 280 & 9                        & 352                             & 641                       & Accuracy  \\ \midrule
Treatment Planning  & SEER  & Recalling facts from \textbf{Pre-training} & 166 & 9                        & 97                             & 272                       & Accuracy  \\ \midrule
Clinical Outcome Prediction  & MIMIC4ED Benchmark  & Recalling facts from \textbf{Pre-training} & 548	& 1	 & 34	& 583                      & Accuracy  \\ \midrule
Text Classification  & HoC  & Recalling facts from \textbf{Pre-training} & 324
& 10	 & 127	& 460
& Accuracy  \\ \midrule
Fact Verification  & PUBHEALTH Explanation, PubMedQA Verication, EBMS  & Recalling facts from \textbf{Pre-training} &  315	& 27	 & 47	&       389          &Accuracy   \\ \midrule
Natural Language Inference  & MedNLI  &Recalling facts from \textbf{Pre-training} &  33	& 5	 & 	 78       & 115 & Accuracy  \\ \midrule
Text Summarization  & MIMIC-IV, RCT, MIMIC-CXR, MedQSum & Retrieving facts from \textbf{Contexts} & 372 & 79                        & 20                             & 470                       & BLEU, ROGUE \\ \midrule
Information Extraction  & ADE, PICO, PMC-Patient & Retrieving facts from \textbf{Contexts} & 137 & 7                        & 126                             & 268                       & BLEU, ROGUE \\ \midrule
Named Entity Recognition  & BC4Chem, BC5Chem, BC5Disease, Species800  & Retrieving facts from \textbf{Contexts} & 35 & 5                        & 52                             & 92                       & Accuracy, Recall, Precision, F1  \\ 
\bottomrule

\end{tabular}
\end{threeparttable}
\end{sidewaystable}

\subsection{Detail Tasks in MedS-Ins}
\label{sec:detail MedSIns}
\footnotesize
In Table~\ref{tab:Detailed instruction tuning dataset}, we further list out the detailed tasks included in MedS-Ins.

\begin{longtable}{p{5.5cm}|p{3cm}|p{3cm}|p{3cm}}
\caption{The detailed 122 tasks in our instruction dataset, along with the corresponding sources, categories, and domains.} 
\label{tab:Detailed instruction tuning dataset} \\
\toprule
Task & Source & Category & Domain \\ \midrule
\endfirsthead
\toprule
Task & Source & Category & Domain \\ \midrule
\endhead

\hline \multicolumn{4}{r}{{Continued on next page}} \\ \hline
\endfoot
\hline \hline
\endlastfoot

        participant extraction & PICO & Information Extraction & Academical Papers \\ 
        intervention extraction & PICO & Information Extraction & Academical Papers \\ 
        outcome extraction & PICO & Information Extraction & Academical Papers \\ 
        pubmedqa question generation & PubMedQA & Text Completion & Academical Papers \\ 
        pubmedqa classification & PubMedQA & Fact Verification & Academical Papers \\ 
        pubmedqa question generation & PubMedQA & Text Completion & Academical Papers \\ 
        pubmedqa classification & PubMedQA & Intent Identification & Academical Papers \\ 
        pubmedqa answer generation & PubMedQA & Question Answering & Academical Papers \\ 
        healthfact classification & HealthFact & Fact Verification & Medical Knowledge \\ 
        healthfact sentence generation & HealthFact & Sentence Composition Analysis & Medical Knowledge \\ 
        healthfact sentence generation & HealthFact & Explanation & Medical Knowledge \\ 
        europa ecdc tm en sv translation & ECDC-TM & Translation & Medical Knowledge \\ 
        europa ecdc tm en de translation & ECDC-TM & Translation & Medical Knowledge \\ 
        europa ecdc tm fr en translation & ECDC-TM & Translation & Medical Knowledge \\ 
        head qa answer generation & HeadQA & Question Answering & Exam \\ 
        head qa language translation en to es & HeadQA & Translation & Exam \\ 
        head qa language translation es to en & HeadQA & Translation & Exam \\ 
        head qa classification & HeadQA & Text Classification & Exam \\ 
        drug extraction ade & ADE Corpus V2 & Named Entity Recognition & Medical Knowledge \\ 
        disease entity extraction ncbi dataset & NCBI Disease Corpus & Named Entity Recognition & Medical Knowledge \\ 
        disease entity extraction bc5cdr dataset & BioCreative V & Named Entity Recognition & Medical Knowledge \\ 
        drug dose extraction & ADE Corpus V2 & Information Extraction & Medical Knowledge \\ 
        gene extraction bc2gm dataset & BC2GM & Named Entity Recognition & Medical Knowledge \\ 
        gene extraction linnaeus dataset & Linnaeus Corpus & Named Entity Recognition & Medical Knowledge \\ 
        cell extraction anem dataset & ANEM & Named Entity Recognition & Medical Knowledge \\ 
        organism substance extraction anem dataset & ANEM & Named Entity Recognition & Medical Knowledge \\ 
        adverse drug event classification & ADE Corpus V2 & Text Classification & Medical Knowledge \\ 
        medical question pair dataset text classification & Medical Question Pair Dataset & Text Retrieval & Daily Conversation \\ 
        dataset card for catalonia independence corpus text classification & Catalonia Independence Corpus & Intent Identification & Medical Knowledge \\ 
        lima medical filtered answer generation & LIMA & Question Answering & Daily Conversation \\ 
        mednli textual entailment discrimitive & MedNLI & Natural Language Inference & Clinical Text \\ 
        mednli textual entailment generative & MedNLI & Natural Language Inference & Clinical Text \\ 
        mednli wrong textual entailment generative & MedNLI & Wrong Candidate Generation & Clinical Text \\ 
        mednli textual entailment classification & MedNLI & Natural Language Inference & Clinical Text \\ 
        do synonyms generation & DO & Text Completion & Medical Knowledge \\ 
        do entity explanation & DO & Explanation & Medical Knowledge \\ 
        do entity relation classification & DO & Word Relation Classification & Medical Knowledge \\ 
        biolord text matching & BioLORD & Text Retrieval & Medical Knowledge \\ 
        biolord summarization & BioLORD & Text Summarization & Medical Knowledge \\ 
        biolord explanation & BioLORD & Explanation & Medical Knowledge \\ 
        mmedbench explanation chinese & MMedBench & Explanation & Medical Knowledge \\ 
        mmedbench explanation english & MMedBench & Explanation & Medical Knowledge \\ 
        mmedbench explanation french & MMedBench & Explanation & Medical Knowledge \\ 
        mmedbench explanation japanese & MMedBench & Explanation & Medical Knowledge \\ 
        mmedbench explanation russian & MMedBench & Explanation & Medical Knowledge \\ 
        mmedbench explanation spanish & MMedBench & Explanation & Medical Knowledge \\ 
        medqa question answering en & MedQA & Question Answering & Medical Knowledge \\ 
        medqa question answering zh & MedQA & Question Answering & Medical Knowledge \\ 
        igakuqa question answering & IgakuQA & Question Answering & Medical Knowledge \\ 
        frenchmedmcqa question answering & FrenchMedMCQA & Question Answering & Medical Knowledge \\ 
        rumedbench question answering & RuMedBench & Question Answering & Medical Knowledge \\ 
        liveqa medical filtered conversation & LiveQA & Dialogue & Daily Conversation \\ 
        medmcqa multiple choice explanation qa & MedMCQA & Explanation & Exam \\ 
        medmcqa multiple choice qa & MedMCQA & Question Answering & Exam \\ 
        medicationqa question answering & MedicationQA & Question Answering & Exam \\ 
        chatdoctor conversation & ChatDoctor & Dialogue & Daily Conversation \\ 
        chatdoctor health care magic conversation & ChatDoctor & Dialogue & Daily Conversation \\ 
        chatdoctor icliniq conversation & ChatDoctor & Dialogue & Daily Conversation \\ 
        medalplaca wikidoc textbook qa & WikiDoc & Question Answering & Medical Knowledge \\ 
        medalplaca wikidoc patient information qa & WikiDoc & Question Answering & Medical Knowledge \\ 
        medalplaca flashcard qa & MedAlpaca & Question Answering & Medical Knowledge \\ 
        report entity extraction & Radiopaedia Caption NER & Sentence Composition Analysis & Clinical Text \\ 
        pmc patient case report title generation & PMC-Patient & Text Completion & Academical Papers \\ 
        pmc patient case report basic information extraction & PMC-Patient & Information Extraction & Academical Papers \\ 
        pmc patient similar case retrieval & PMC-Patient & Text Retrieval & Academical Papers \\ 
        mimic iv note discharge instruction completion & MIMIC-IV-Notes & Text Completion & Medical Knowledge \\ 
        clinical trials gov data named entity recognition & Clinical-Trials.gov & Named Entity Recognition & Exam \\ 
        mimic ultrasound summarization & MIMIC-IV & Text Summarization & Clinical Text \\ 
        mimic angiography summarization & MIMIC-IV & Text Summarization & Clinical Text \\ 
        mimic mammogram summarization & MIMIC-IV & Text Summarization & Clinical Text \\ 
        mimic pathology summarization & MIMIC-IV & Text Summarization & Clinical Text \\ 
        mimic fluoroscopy summarization & MIMIC-IV & Text Summarization & Clinical Text \\ 
        mimic ct chest summarization & MIMIC-IV & Text Summarization & Clinical Text \\ 
        mimic ct head and neck summarization & MIMIC-IV & Text Summarization & Clinical Text \\ 
        mimic ct brain summarization & MIMIC-IV & Text Summarization & Clinical Text \\ 
        mimic ct abdomen summarization & MIMIC-IV & Text Summarization & Clinical Text \\ 
        mimic ct pelvis summarization & MIMIC-IV & Text Summarization & Clinical Text \\ 
        mimic ct spine summarization & MIMIC-IV & Text Summarization & Clinical Text \\ 
        mimic mri chest summarization & MIMIC-IV & Text Summarization & Clinical Text \\ 
        mimic mri head and neck summarization & MIMIC-IV & Text Summarization & Clinical Text \\ 
        mimic mri brain summarization & MIMIC-IV & Text Summarization & Clinical Text \\ 
        mimic mri abdomen summarization & MIMIC-IV & Text Summarization & Clinical Text \\ 
        mimic mri pelvis summarization & MIMIC-IV & Text Summarization & Clinical Text \\ 
        mimic mri spine summarization & MIMIC-IV & Text Summarization & Clinical Text \\ 
        mimic modality classification & MIMIC-IV & Text Classification & Clinical Text \\ 
        mimic anatomy classification & MIMIC-IV & Text Classification & Clinical Text \\ 
        medlineplus drug qa & MedlinePlus & Question Answering & Medical Knowledge \\ 
        medlineplus encyclopedia qa & MedlinePlus & Text Summarization & Medical Knowledge \\ 
        ebms question answering & EBMSummariserCorpus & Question Answering & Medical Knowledge \\ 
        ebms answer vertification & EBMSummariserCorpus & Question Answering & Medical Knowledge \\ 
        mashqa question answering & MASHQA & Question Answering & Medical Knowledge \\ 
        mediqa ans question answering & MEDIQA-AnS & Question Answering & Medical Knowledge \\ 
        medquad question answering & MedQuAD & Question Answering & Medical Knowledge \\ 
        bioasq question answering & BIOASQ & Question Answering & Medical Knowledge \\ 
        covidqa question answering & COVID-QA & Question Answering & Medical Knowledge \\ 
        hoc text classification & HoC & Text Classification & Medical Knowledge \\ 
        pubmedmesh text classification & PubMed MeSH & Text Classification & Medical Knowledge \\ 
        litcovid text classification & LitCovid & Text Classification & Medical Knowledge \\ 
        ms2 text summurization & MS2 & Text Summarization & Medical Knowledge \\ 
        rct text summurization & RCT & Text Summarization & Medical Knowledge \\ 
        pubmed text summarization & PubMed & Text Summarization & Medical Knowledge \\ 
        mimic cxr text summurization & MIMIC-CXR & Text Summarization & Clinical Text \\ 
        readibility text summurization & Readibility & Text Summarization & Medical Knowledge \\ 
        medqsum text summurization & MeQSum & Text Summarization & Medical Knowledge \\ 
        icd10 code explanation & ICD-10-CM & Explanation & Medical Knowledge \\ 
        mimic4 icd10 text clissification & MIMIC4 ICD-10 & Text Classification & Clinical Text \\ 
        mimic4ed benchmark hospitalization & MIMIC4ED Benchmark & Clinical Outcome Prediction & Clinical Text \\ 
        mimic4ed 72h ed revisit disposition & MIMIC4ED Benchmark & Clinical Outcome Prediction & Clinical Text \\ 
        mimic4ed benchmark critical triage & MIMIC4ED Benchmark & Clinical Outcome Prediction & Clinical Text \\ 
        cmedqa2 question answering & CMedQA2 & Question Answering & Medical Knowledge \\ 
        pubmedqa train set & PubMedQA & Question Answering & Academical Papers \\ 
        bc4chem named enetity recognition & BC4-Chem & Named Entity Recognition & Medical Knowledge \\ 
        bc5chem named enetity recognition & BC5-Chem & Named Entity Recognition & Medical Knowledge \\ 
        species800 named enetity recognition & Species-800 & Named Entity Recognition & Medical Knowledge \\ 
        headqa question answering & HeadQA & Question Answering & Medical Knowledge \\ 
        DDXPlus text classification train & DDXPlus & Diagnosis & Medical Knowledge \\ 
        SEER text classification train & SEER & Treatment Planning & Medical Knowledge \\ 
        fact verication short medical test train & MedlinePlus & Fact Verification & Medical Knowledge \\ 
        fact verication short genetic conditions train & MedlinePlus & Fact Verification & Medical Knowledge \\ 
        fact verication short gene train & MedlinePlus & Fact Verification & Medical Knowledge \\ 
        fact verication short encyclopedia train & MedlinePlus & Fact Verification & Medical Knowledge \\ 
        fact verication short drugs train & MedlinePlus & Fact Verification & Medical Knowledge \\
        \bottomrule

\end{longtable}

\end{document}